\documentclass[jair,twoside,11pt,theapa]{article}
\usepackage{jair, theapa, rawfonts}

\usepackage{subfigure}

\usepackage{threeparttable}

\usepackage{amsmath,amssymb}
\usepackage{color}
\usepackage{colortbl}
\definecolor{Gray}{gray}{0.92}
\usepackage[first=0,last=9]{lcg}
\usepackage{multirow}
\usepackage{comment}
\usepackage{tabularx}
\usepackage{footnote}

\newcolumntype{v}{>{\hsize=.35\hsize}X}
\newcolumntype{d}{>{\hsize=.65\hsize}X}

\newcolumntype{V}{>{\hsize=.25\hsize}X}
\newcolumntype{B}{>{\hsize=1.2\hsize}X}
\newcolumntype{S}{>{\hsize=.55\hsize}X}

\newcolumntype{a}{>{\hsize=.11\hsize}X}
\newcolumntype{n}{>{\hsize=.5\hsize}X}

\usepackage{algorithm}
\usepackage[noend]{algpseudocode}
\usepackage{graphicx}

\usepackage{xr}

\newtheorem{defn}{Definition}

\def \eg {\emph{e.g.} }
\def \Eg {\emph{E.g.} }
\def \ie {\emph{i.e.} }

\def \depththreshold {$\textit{thresh}_\textit{convex}$ }

\def \vertices [#1]{$V_{#1}$}
\def \subgraphvertices [#1][#2]{$V^{'#1}_{#2}$}
\def \subgraphmultivertices [#1][#2][#3]{$V^{'#1,#2}_{#3}$}

\def \eqsubgraphvertices [#1][#2] {V^{'#1}_{#2}}
\def \eqsubgraphmultivertices [#1][#2][#3] {V^{'#1,#2}_{#3}}

\def \maximumdepth {$\textit{dmax}$ }
\def \minimumdepth {$\textit{dmin}$ }
\def \maxconcave {$d_c\textit{max}$ }
\def \minconcave {$d_c\textit{min}$ }

\jairheading{N}{2023}{S-E}{08/21}{03/23}
\ShortHeadings{Object-agnostic Affordance Categorization via Unsupervised Learning}
{Toumpa \& Cohn}
\firstpageno{1}

\begin{document}

\title{Object-agnostic Affordance Categorization via Unsupervised Learning of Graph Embeddings}
       
\author{\name Alexia Toumpa \email A.Toumpa@leeds.ac.uk \\
       \addr School of Computing, University of Leeds, UK
       \AND
       \name Anthony G. Cohn \email A.G.Cohn@leeds.ac.uk \\
       \addr School of Computing, University of Leeds, UK \\
       Tongji University, China}


\maketitle

\begin{abstract}
Acquiring knowledge about object interactions and affordances can facilitate scene understanding and human-robot collaboration tasks.
As humans tend to use objects in many different ways depending on the scene and the objects' availability, learning object affordances in everyday-life scenarios is a challenging task, particularly in the presence of an open set of interactions and objects.
We address the problem of affordance categorization for class-agnostic objects with an open set of interactions; we achieve this by learning similarities between object interactions in an unsupervised way and thus inducing clusters of object affordances.
A novel depth-informed qualitative spatial representation is proposed for the construction of {\em Activity Graphs} (AGs), which abstract from the continuous representation of spatio-temporal interactions in RGB-D videos.
These AGs are clustered to obtain groups of objects with similar affordances.
Our experiments in a real-world scenario demonstrate that our method learns to create object affordance clusters with a high V-measure even in cluttered scenes. The proposed approach handles object occlusions by capturing effectively possible interactions and without imposing any object or scene constraints.

\end{abstract}


\section{Introduction} \label{sec:introduction}
In the literature, the meaning of the term \textit{affordance} of an object differs depending on the context. In robotic applications, \eg robot manipulation tasks, the definition of \textit{affordance} is bound to the part of a tool, which can be afforded in a specific way, \eg the handle of a hammer has the affordance of `hold' whereas the head has the affordance of `hit', the inner surface of a cup has the affordance of `contain' although the cup's handle has the affordance of `hold'.
In contrast, in human-object interaction recognition tasks, \textit{affordance} is defined as the way an object can be utilized by the human in a scene, \eg if a human uses a cup for containing something then the cup will have the affordance of `contain'.
Moreover, an object may have more than one affordance as it depends on the purpose it is being used for, \eg a pizza box can have the affordance of `contain' when it is being utilized as a container of a pizza or `support' when is plays the role of a tray. Such multi-labeled affording objects can be recognized by considering their interactions with other objects.

In a human-robot collaboration scenario, acquiring knowledge of the affordances of the objects in a scene is crucial for aiding the human, \eg assisting the human when performing a physically hard task. This becomes challenging when the scene comprises an open-set of objects and the affordance space enlarges. Moreover, in human action prediction tasks, the affordances of the objects carry useful information for the prediction of the future action, and are highly correlated with the rest of the objects the human interacts with. Nevertheless, such knowledge is not easy to obtain as humans tend to use the same object in different ways depending on the task, thus changing its primary affordance.

Object \emph{affordances} were first formally defined by Gibson \cite{gibson1977theory}; however the concept of object affordances as it is understood in Computer Science has not been explicitly defined in the literature, causing some confusion. For this purpose we propose the definition:
\begin{defn} \label{def:affordance}
An affordance is a property of an object arising from its interaction with another entity, i.e. agent, object. It is correlated with the occurring interaction as every interaction exploits at least one object affordance. An object may have multiple affordances based on the various interactions it may have with other entities.
\end{defn}

Based on this definition, we address the problem of affordance inference by exploiting object interactions through RGB-D video data. Hence, we rely on the way objects are being utilized by the human agents in a scene, allowing objects to support different kinds of affordances at the same time.
Thus, we focus on learning groups of high-level object interactions, which take into account their spatio-temporal relations from extracted visual appearances. 
Graphs are able to represent both static and dynamic high level relations between objects.
Accordingly, we embody these object interactions through a high-level graphical structure, the {\em Activity Graph}, abstracting from the continuous spatio-temporal representation and acquiring depth-informed qualitative spatial relations between object pairs.
With these high-level qualitative relations, our graphical structures for capturing interactions are not dependant on scene-specific characteristics, \ie the objects' labels, distance between objects, etc.

Definition \ref{def:affordance} states that
{\em ``every interaction exploits at least one object affordance''}; affordances of objects are inferred from these high-level graphs representing pair-wise object interactions. Affordance clusters are formed in an unsupervised way by exploiting graph similarity using the cosine similarity measure of their embeddings in a learned latent space. By clustering graph structures, a hierarchical tree representation is produced demonstrating their similarity. Since our approach is based on learning a high-level representation of interactions, it is not limited to any number or kind of affordances, scenes, and objects.

To obtain a richer set of spatial relationships than those possible from a sequence of purely 2D frames, we exploit the depth information assuming the presence of RGB-D video data. The depth cues allow some inference about the morphology of the objects in the scene and thus the possible ways they can interact with other objects, 
\eg non-concave objects can not act as \emph{containers}.

Hence, the objectives of this work are \footnote{Preliminary versions of this work in unpublished workshop papers can be found in \cite{toumpa2019relational} and \cite{toumpadepth}.}:
\begin{itemize}
    \item to propose a depth-informed set of qualitative relations, which can describe a wide variety of object configurations regarding a real-world scenario as well as differentiating between object occlusions and interactions, for detecting effectively spatial relationships of objects;
    \item to introduce a novel unsupervised affordance categorization framework, which handles an open-set of interactions and affordances, considering high-level information of the human-object and object-object interactions;
    \item to present a novel human-object interaction dataset, which focuses on different object affordances.
\end{itemize}


\section{Literature Review} \label{sec:literature_review}

\noindent \textbf{Qualitative spatial relations}
Qualitative relations are mostly used to represent semantically meaningful properties of a perceived \emph{event}, \ie interaction, deformation, whilst abstracting away from any non-relevant information, \ie pixel values, numeric measurements. Spatial relation formalisms comprise: mereotopology, direction, distance, shape, relationships between static spatial entities, and moving objects \fullcite{chen2015survey}, \fullcite{dylla2017survey}, \fullcite{landsiedel2017review}.
Due to the nature of the problem we address in this work, we focus on mereotopological information for representing human-object and object-object interactions and shape information for defining possible concave regions on the detected objects. 

The Region Connection Calculus (RCC) \fullcite{cohn1997qualitative}, \fullcite{gerevini2002combining} and the n-intersection model \fullcite{egenhofer2005spherical}, \fullcite{egenhofer1991point}, \fullcite{egenhofer1995equivalence}, \fullcite{egenhofer1990categorizing}, \fullcite{egenhofer1993topological}, \fullcite{egenhofer2007spatial}, \fullcite{egenhofer1994topological}, \fullcite{egenhofer19949} are the most well-known and well-studied approaches for representing topological information. However, as both sets of relations only rely on the spatial location of the detected entities, \ie objects, no occlusion can be differentiated from an actual interaction of containment or support.
Galton's \emph{Lines of Sight} \fullcite{galton1994lines}, the \emph{Region Occlusion Calculus} \fullcite{randell2001images}, and \emph{The Occlusion Calculus} (OCC) \fullcite{kohler2002occlusion} aim to solve the occlusion problem in 2D space. However, relations as `contain' and `support' are still not detectable.
Thus, shape characteristics are crucial for detecting more relevant relations to object interactions.

Qualitative shape information can be obtained by using region-based methods \fullcite{cohn1995hierarchical} or boundary-based methods \fullcite{leyton1988process}, \fullcite{galton1999qualitative}, \fullcite{gottfried2003atripartite}, \fullcite{gottfried2003btripartite}, \fullcite{gottfried2004reasoning}. Both kind of methods depend on the 2D shapes of the detected entities. By exploiting the depth cues of the entities, concave regions are discovered in the 3D space, indicating possible containment.

Inspired by the RCC8 spatial relations and the \textit{Process-Grammar} \fullcite{leyton1988process}, in this work we propose a novel set of mereogeometrical relations that exploit the concave regions indicated by the depth cues of the detected entities. Our new set of relations define more complex and concave-informed relations, such as `contain' and `support'.

\setlength{\tabcolsep}{4pt}
\begin{table}
\centering
\begin{threeparttable}
\caption{Related works on object affordance detection.}
\label{table:literature}
\begin{tabular}{c|c|cccc|ccc}
\hline
\noalign{\smallskip}
& Input data type & \multicolumn{4}{c}{Information exploited \tnote{a}} & \multicolumn{3}{c}{Solve task\tnote{b}}\\
& & L.F. & H.A. & H.F. & O.T & O.O. & O.S A. & O.S.O\\
\noalign{\smallskip}
\noalign{\smallskip}
\rowcolor{Gray}
\small\shortcite{montesano2009learning} & 2D \& Keypoints & \checkmark & & & & & & \\
\small\shortcite{zhao2013scene} & 2D \& 3D & \checkmark & & & & & & \\
\rowcolor{Gray}
\small\shortcite{myers2015affordance} & 2.5D & \checkmark & & & & \checkmark & & \\
\small\shortcite{nguyen2016detecting} & 2.5D & \checkmark & & & & & & \\
\rowcolor{Gray}
\small\shortcite{nguyen2017object} & 2D & \checkmark & & & & & & \\
\small\shortcite{kokic2017affordance} & Synthetic & \checkmark & & & & \checkmark & & \\
\rowcolor{Gray}
\small\shortcite{sawatzky2017weakly} & 2D \& Keypoints & \checkmark & & & & & & \\
\small\shortcite{do2018affordancenet} & 2D & \checkmark & & & & & & \\
\rowcolor{Gray}
\small\shortcite{wang2020learning} & 2D & \checkmark & & & & & & \\
\small\shortcite{deng20213d} & 3D & \checkmark & & & & \checkmark & & \\
\rowcolor{Gray}
\small\shortcite{xu2021affordance} & 2.5D \& Keypoints & \checkmark & & & & \checkmark & & \\

\small\shortcite{turek2010unsupervised} & 2D & & & & \checkmark & & \checkmark & \checkmark \\
\rowcolor{Gray}
\small\shortcite{qi2018human} & 3D & & & \checkmark & & & & \\

\small\shortcite{kjellstrom2011visual} & 2D & & \checkmark & & & \checkmark & \checkmark & \\
\rowcolor{Gray}
\small\shortcite{yao2013discovering} & 3D & & \checkmark & & & & & \\
\small\shortcite{qi2017predicting} & 2.5D & & \checkmark & \checkmark & & & & \\
\rowcolor{Gray}
\small\shortcite{gkioxari2018detecting} & 2D & & \checkmark & & & & & \\
\small\shortcite{fang2018demo2vec} & 2D & & & \checkmark & & & & \\
\rowcolor{Gray}
\small\shortcite{chuang2018learning} & 2D & & \checkmark & \checkmark & & & & \\
\small\shortcite{tan2019object} & 2D & & \checkmark & \checkmark & & & & \\
\rowcolor{Gray}
\small\shortcite{wu2020chair} & Synthetic & & \checkmark & & & & & \\
\small\shortcite{hou2021affordance} & 2D & & \checkmark & & & & & \\

\rowcolor{Gray}
\small\shortcite{sridhar2008learning} & 2D & & & \checkmark & & & \checkmark & \checkmark \\
\small\shortcite{aksoy2010categorizing} & 2D & & & \checkmark & & & \checkmark & \\
\rowcolor{Gray}
\small\shortcite{aksoy2011learning} & 2D & & & \checkmark & & & \checkmark & \\
\small\shortcite{pieropan2013functional} & 2.5D & & & \checkmark & & & \checkmark & \\
\rowcolor{Gray}
\small\shortcite{pieropan2014recognizing} & 2D & & & \checkmark & \checkmark & & & \\
\small\shortcite{moldovan2014occluded} & Synthetic & & & & & \checkmark & & \\
\rowcolor{Gray}
\small\shortcite{liang2016inferring} & 2.5D & & & \checkmark & \checkmark & \checkmark & & \\
\small\shortcite{liang2018tracking} & 2.5D & & & & \checkmark & \checkmark & & \\
\rowcolor{Gray}
This work & 2.5D & & & \checkmark & & \checkmark & \checkmark & \checkmark \\
\hline
\end{tabular}
\begin{tablenotes}
    \item[a] L.F.: Low-level features\hspace*{1cm}H.A.: Human Actions\hspace*{1cm}H.F.: High-level Features (graph representations, embeddings)\hspace*{1cm}O.T.: Object tracks 
    \item[b] O.O: Object Occlusions\hspace*{1cm}O.S.A.: Open-set of affordances\hspace*{1cm}O.S.O: Open-set of objects
\end{tablenotes}
\end{threeparttable}
\end{table}
\setlength{\tabcolsep}{1.4pt}

\noindent \textbf{Object affordances}
Several methods have been proposed for detecting functional object parts and their corresponding affordance labels. These works involve the detection of object affordance parts by considering their visual and geometric features \fullcite{deng20213d}, \fullcite{wang2020learning}. One of the early works in this direction focused in the detection of graspable object areas by creating local visual descriptors of grasping points and estimating the probability of the presence of a graspable object based on the Bernoulli trial \fullcite{montesano2009learning}. New approaches employ Convolutional Neural Network (CNN) models to produce classes of functional object parts from RGB \fullcite{nguyen2017object}, \fullcite{do2018affordancenet}, \fullcite{sawatzky2017weakly} and synthetic data \fullcite{kokic2017affordance}. However, depth cues along with the RGB information have demonstrated a greater detection accuracy in this task \fullcite{nguyen2016detecting}, \fullcite{myers2015affordance}, \fullcite{xu2021affordance}.
Additionally, incorporating knowledge about the scene and context in which an object is being used boosts the prediction accuracy even more \fullcite{zhao2013scene}.

However, processing static visual information restricts the number of affordances assigned to an object to be the ones correlated only with its visual features.
For this purpose, many works have considered exploiting the correlation of human actions and the detected objects in a scene \fullcite{gkioxari2018detecting}, \fullcite{yao2013discovering}, \fullcite{fang2018demo2vec}, \fullcite{kjellstrom2011visual}, \fullcite{hou2021affordance}, \fullcite{wu2020chair}. Depending on the human-object interaction being held, a different affordance is detected.
Also, prediction of object affordances, cast as human-object interactions, utilize graph-based convolutional neural networks, which are trained on Object Affordance Graphs (OAG), constructed by the spatio-temporal relations between human and objects/scenes across the input video data \fullcite{tan2019object}.
Similarly, a spatio-temporal And-Or graph (ST-AOG) exploits the context captured by objects, actions, and affordances to solve the task of activity understanding \fullcite{qi2017predicting}.
Another approach predicts action-object affordances based on the contextual information of the scene, whilst deploying a graph to represent the objects in the scene as the nodes and their spatial relations as the edges \fullcite{chuang2018learning}.
These works demonstrate that by fusing knowledge about the scenario in which an interaction takes place facilitates the prediction of affordances, however limits the generalizability across different domains.

To facilitate domain independence, high-level graph representations of interactions are sometimes employed.
Recent works introduce such graphical structures in synthetic indoor environments focusing on the prediction of furniture areas the human is most likely to interact with \fullcite{qi2018human}, whereas outdoors scenes are examined by considering the behavior of moving objects around them \fullcite{turek2010unsupervised}.
Nevertheless, these approaches only consider a one-to-one mapping of affordances and objects, hence the objects are bound to a single kind of interaction, not permitting multi-labeled object affordances.

Though graphical structures are able to extract high-level information about the interactions in the scene, their structure might be the cause of domain restriction \fullcite{aksoy2010categorizing}, \fullcite{aksoy2011learning}, \fullcite{pieropan2014recognizing}, \fullcite{pieropan2013functional}. For this purpose, qualitative spatio-temporal relations are exploited for their construction \fullcite{sridhar2008learning}.
One of the fundamental obstacles in these works is object occlusion.
To mitigate this problem, the tracks and visual appearances and disappearances of non-deformable objects are considered  \fullcite{liang2018tracking}, \fullcite{liang2016inferring}, \fullcite{moldovan2014occluded}. However, these approaches are restricted to the detection of a single object affordance, \ie containment, and prone to false positive detections of the containment relation due to occlusions.
Table \ref{table:literature} presents the related works of object affordance detection and prediction.


In this work, we propose a novel method for predicting object affordances in real-world scenarios with object occlusions.
Different from any other published work, our approach is not restricted to a predefined set of objects, interactions, or scenes. High-level graphs assist in considering an open-set of interactions and no object labels restrict the generalizability of our method. Our experiments demonstrate that, without any supervision, we acquire homogeneous and complete clusters of object affordances by exploiting qualitative information about their interactions and their shapes.



\section{Background} \label{sec:background}
\subsection{Activity Graphs} \label{sec:activity_graphs}

Relational graph structures represent high-level information by abstracting from the continuous space of the exploited relations.
From definition \ref{def:affordance}: \textit{``It (an affordance) is correlated to the occurring interaction as every interaction exploits at least one object affordance.''}, hence an object-object and human-object interaction reveals each object's affordances. Relational graph structures of object interactions aid in representing an open set of interactions. 
Thus, in this work we exploit the representation of \emph{Activity Graphs} ($AGs$) \cite{sridhar2010relational}, which are relational graphs that describe the interaction between entities, \ie objects and human body parts, of a scene by considering their qualitative spatio-temporal relationships.

An $AG$ consists of three layers of vertices, where each layer comprises a single type of node and only nodes in adjacent layers can be connected with each other.
The three layers of an $AG$ are: 1) the \emph{entity} layer, which contains the set of vertices of the entities which interact (\vertices[ent] ), 2) the \emph{spatial} layer, which consists of vertices with the spatial relations (\vertices[spat] ) which describe the spatial interactions of the entities in \vertices[ent], and 3) the \emph{temporal} layer, which specifies the temporal relations between the \emph{spatial} layer nodes (\vertices[temp] ).

\begin{table}
\renewcommand{\arraystretch}{1.2}
\centering
\begin{threeparttable}
\caption{RCC2} \label{table:rcc2}
\begin{tabularx}{\linewidth}{vXd}
    \rowcolor{Gray}
    \textbf{RCC2} & \textbf{Definition} & \textbf{Description}\\
    \hline
    \textsf{C}($\alpha$,$\beta$) & $mask(\alpha) \cap mask(\beta) \neq \varnothing $ \tnote{\textdagger} & $\alpha$ and $\beta$ are connected\\
    \rowcolor{Gray}
    \textsf{DC}($\alpha$,$\beta$) & $\lnot \textsf{C}(\alpha,\beta)$ & $\alpha$ is disconnected from $\beta$\\
\end{tabularx}
\begin{tablenotes}
\item[\textdagger] $mask(z)$ defines object $z$'s region in the image plane.
\end{tablenotes}
\end{threeparttable}
\end{table}

\begin{figure}
    \centering
    \includegraphics[trim ={0mm 31mm 0mm 0mm},clip, width=1.\linewidth]{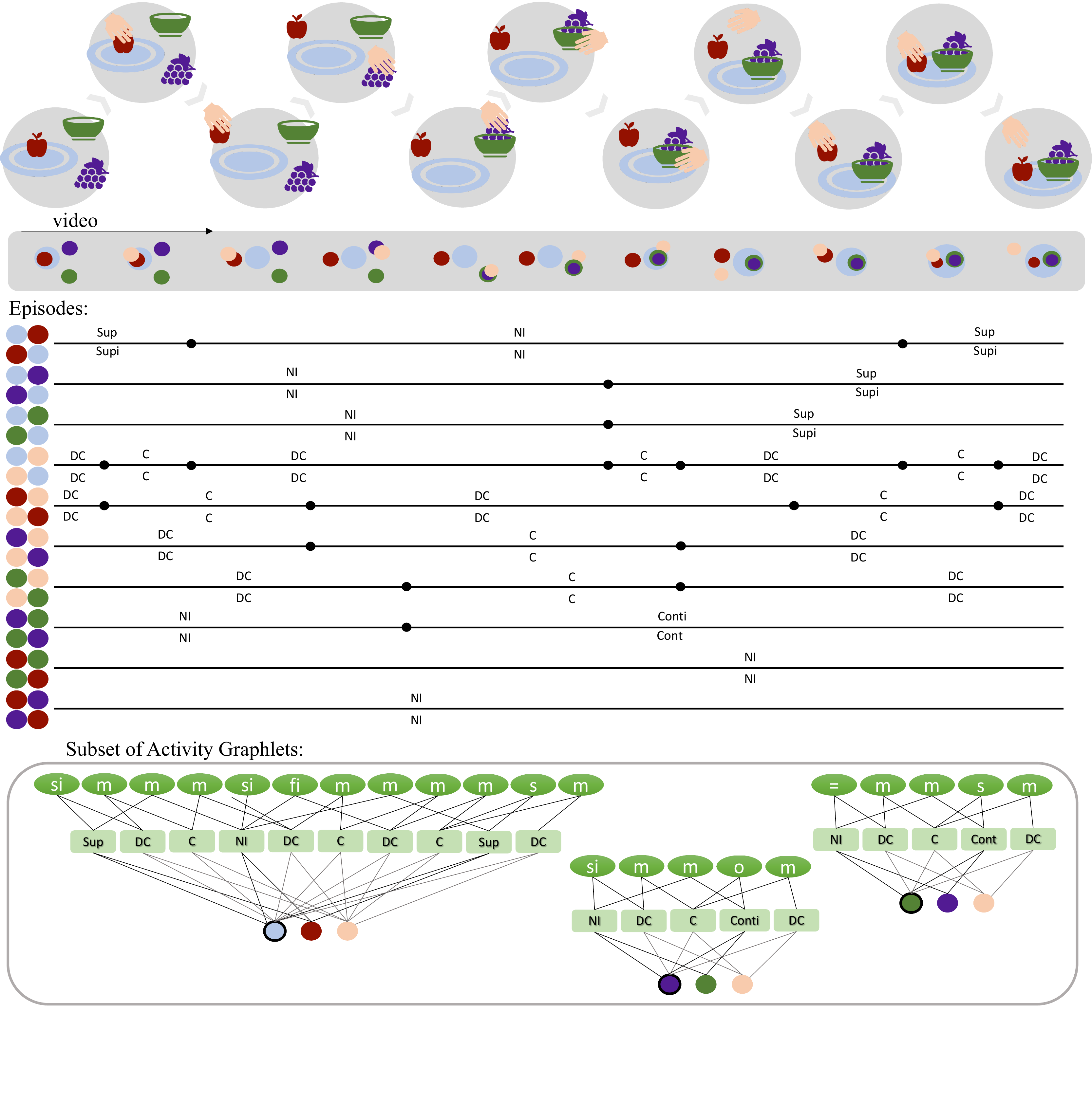}
    \caption{(best viewed in color) Qualitative spatial relations are extracted from detected episodes in the temporal domain of a video. {\em Activity Graphlets} are constructed using these qualitative spatio-temporal relations for individual detected objects (encircled) in the scene describing their interaction with another object and a human body part, \eg human hand.}
    \label{fig:activity_graphs}
\end{figure}

In this work, $AGs$ are deployed to captured object-object and human-object interactions, thus two different qualitative spatial relations (QSRs) are considered, one for each kind of interaction. A set of novel QSRs, introduced in \S \ref{sec:depth_informed_qsr}, is employed to describe the spatial relationships between objects, whereas the RCC2 set of relations (Table \ref{table:rcc2}) \cite{randell1992spatial}, \cite{cohn1997qualitative} are used for representing interactions between objects and humans.
These two spatial relational sets comprise the \vertices[spat] layer of our $AGs$, whilst \emph{Allen's temporal algebra} \cite{allen1983maintaining} is exploited to express the temporal relationships between the spatial relations (\vertices[temp] ).
This qualitative temporal set consist of the relations\footnote{Inverse relations end in \textsf{i}.}: `before' ($<$, $>$), `meets' ({\small\textsf{m}}, {\small\textsf{mi}}), `overlaps', ({\small\textsf{o}}, {\small\textsf{oi}}), `starts' ({\small\textsf{s}}, {\small\textsf{si}}), `during' ({\small\textsf{d}}, {\small\textsf{di}}), `finishes' ({\small\textsf{f}}, {\small\textsf{fi}}), and `equals' ($=$).
The set of qualitative spatial relations for every interaction is obtained from the presence of \emph{episodes}. An \emph{episode} is the maximum period of time where a single spatial relation between two entities holds, while a different spatial relation occurs before and after the defined time period. By considering \emph{episodes} instead of individual frames for detecting the spatial relationships of the objects, our method can effectively generalize across different video fps and is not video length dependent.

We define an \emph{Activity Graphlet} (\emph{AGraphlet}) of an object as a sub-graph of an $AG$, which carries the spatial and temporal information (\subgraphvertices[][spat] ,\subgraphvertices[][temp] ) of the respective object's interactions with another object and a human body part (\subgraphvertices[][obj] ). Figure \ref{fig:activity_graphs} illustrates an example where \emph{AGraphlets} are extracted from a video sequence of three interacting entities.

\subsection{Graph Embedding} \label{sec:graph_embedding}
Graph embeddings are \emph{d}-dimensional vector representations of projections of graphs in a \emph{d}-dimensional latent space. By projecting graphs in a multi-dimensional space we are able to find similarities/dissimilarities between them by considering distance measures on their vector representations.

\begin{figure}
    \centering
    \includegraphics[trim ={20mm 54mm 125mm 12mm},clip, width=.7\linewidth]{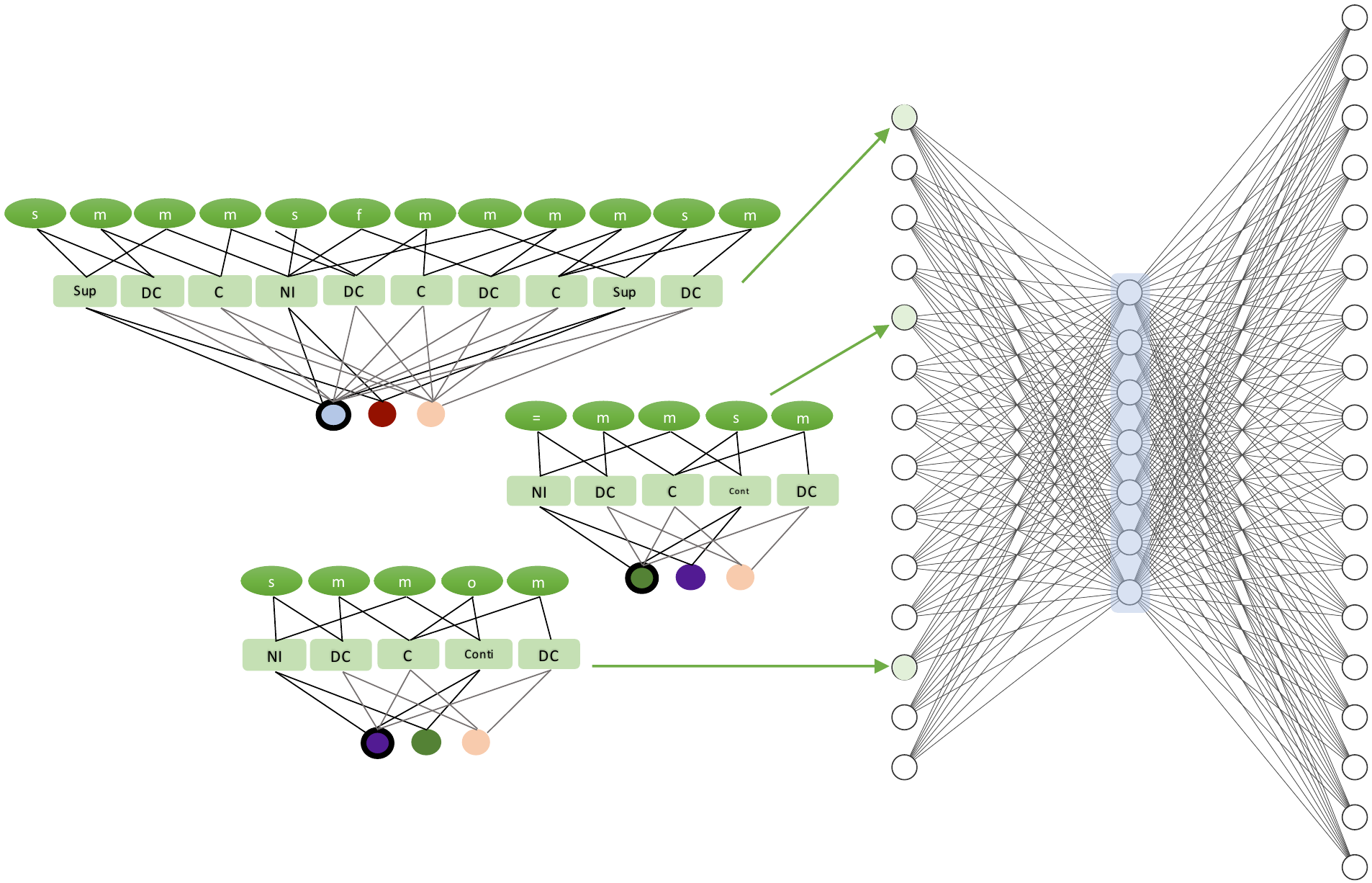}
    \put(-80,160){\footnotesize{Embedding}}
    \put(-120,200){\footnotesize{One-hot}}
    \put(-132,190){\footnotesize{representation}}
    \put(-120,180){\footnotesize{of graphs}}
    \put(-30,210){\footnotesize{Vocabulary}}
    \put(-32,200){\footnotesize{of subgraphs}}
    \put(-110,-10){\footnotesize{Input}}
    \put(-130,-20){\footnotesize{layer $\in \mathbb{R}^{15000}$}}
    \put(-70,10){\footnotesize{Hidden}}
    \put(-80,0){\footnotesize{layer $\in \mathbb{R}^{128}$}}
    \put(-10,-10){\footnotesize{Output}}
    \put(-30,-20){\footnotesize{layer $\in \mathbb{R}^{200000}$}}
    \caption{\emph{Graph2vec} network; the size of the depicted network is specific to one of the folds used for training.}
    \label{fig:graph2vec}
\end{figure}
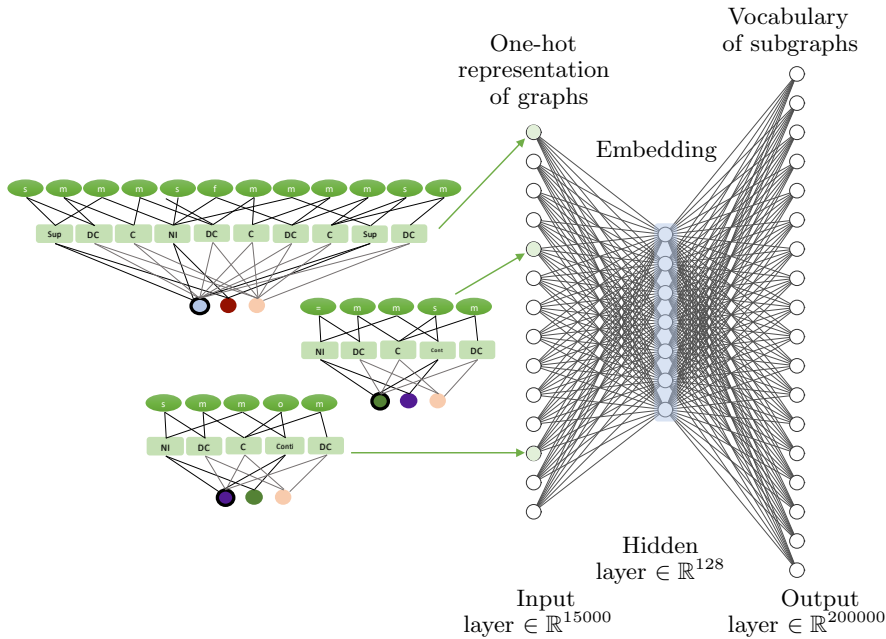

Inspired by the Natural Language Processing literature, we exploit the \emph{graph2vec} network \cite{narayanan2017graph2vec} to learn graph embeddings in an unsupervised way and project the \emph{AGraphlets} in the learned latent space.
The network exploits the notion of context, where context is defined as a fixed number of subgraphs comprising every \emph{AGraphlet}. \emph{AGraphlets} with similar subgraphs represent similar affordances. 
Figure \ref{fig:graph2vec} shows an illustration of the \emph{graph2vec} network with the \emph{AGraphlets} as input data.

We refer the reader to \S \ref{sec:graph2vec_network} for more details about the hyper-parameter's selection and training of the \emph{graph2vec} network.

\section{Overview} \label{sec:overview}
\begin{figure}
    \centering
    \includegraphics[trim ={0mm 52mm 8mm 0mm},clip, width=1.\linewidth]{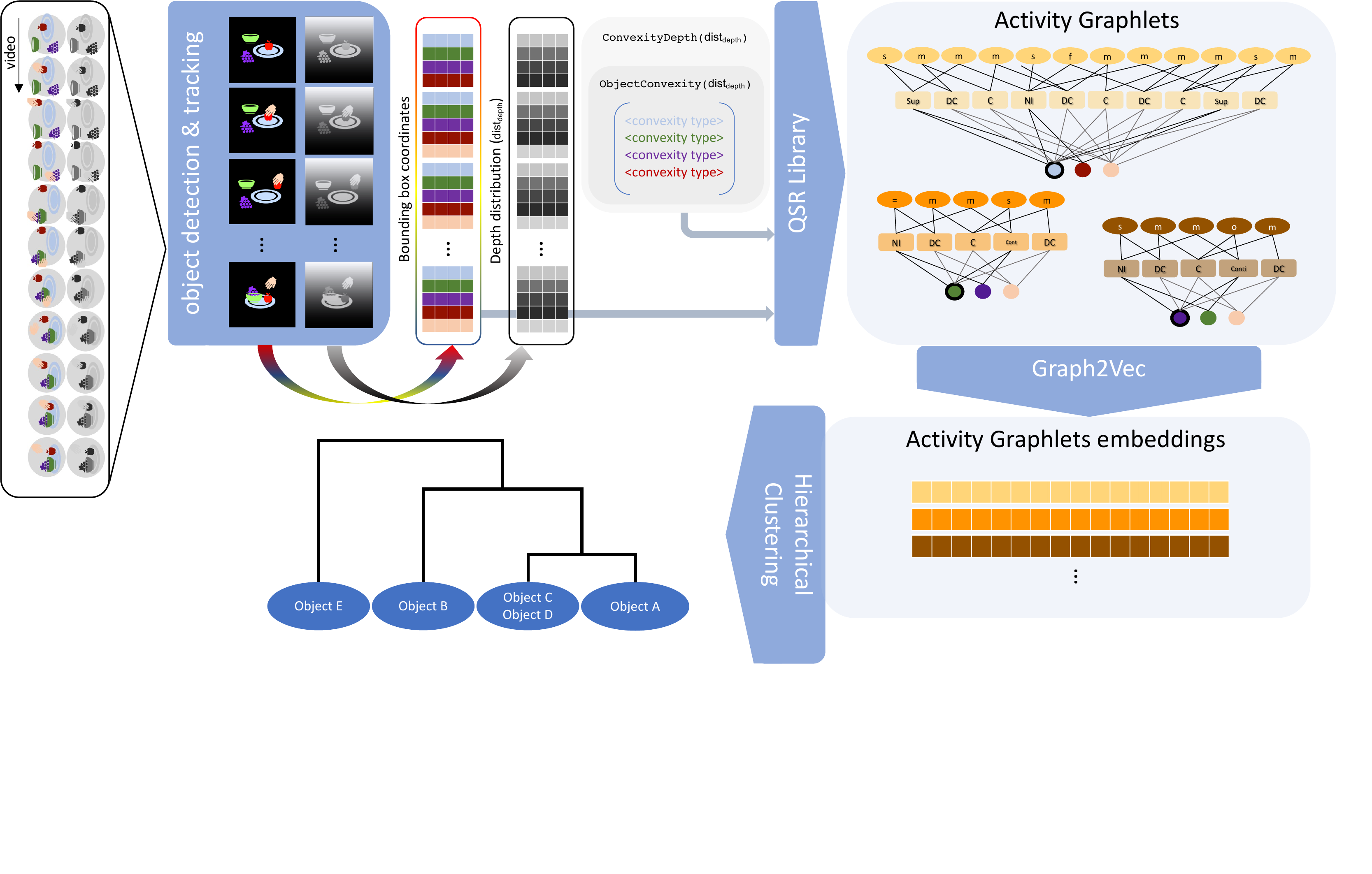}
    \caption{Overview of the proposed approach. Object coordinates and their convexity type are extracted from RGB-D video data per frame and employed for the construction of \emph{AGraphlets} capturing the proposed set of relations. The Graph2Vec network is trained to project these graphs on a latent space. Graph embeddings of \emph{AGraphlets} are then hierarchicaly clustered to create groups of objects with similar affordances.}
    \label{fig:overview}
\end{figure}

The proposed approach solves the problem of object affordance categorization in an unsupervised way, by exploiting depth-informed qualitative information and representing object interactions in a high-level latent space. Figure \ref{fig:overview} summarizes the proposed approach.

Firstly, objects of interest are localized in the input frames of a video by utilizing any visual-based object detector which provides class-agnostic object bounding boxes.
We capture object-object and human-object interactions and we effectively represent qualitative spatial relationships between them by exploiting a novel set of depth-informed spatial relationships (\S \ref{sec:depth_informed_qsr}) and the RCC2 relation set. Such qualitative spatial information constitutes the spatial layer of an \emph{AGraphlet}.
\emph{AGraphlets} are used as a high-level representation of object interactions, thus achieving a generalization across various scenarios. These graphical structures are then employed to train the \emph{graph2vec} network, which projects each \emph{AGraphlet} to a latent space.
Hierarchical clustering, which is described in \S \ref{sec:learning_object_affordances}, is employed for learning in an unsupervised way clusters of detected object interactions.
By exploiting graph embeddings, which are projections of \emph{AGraphlets} onto the latent space, the clustering produces a dendrogram of object affordances.
As, our method is not object-dependent and not scene-constrained, we perform evaluation on the newly presented Leeds Object Affordance Dataset (LOAD), where the groundtruth affordances do not correlate to the objects' primary function.

\section{Depth-informed QSR} \label{sec:depth_informed_qsr}

\subsection{Formulation of Depth-informed Spatial Relations (\emph{DiSR})} \label{sec:formulation_of_disr}

\begin{figure}
    \centering
    \includegraphics[trim ={11mm 126mm 170mm 18mm},clip, width=0.9\linewidth]{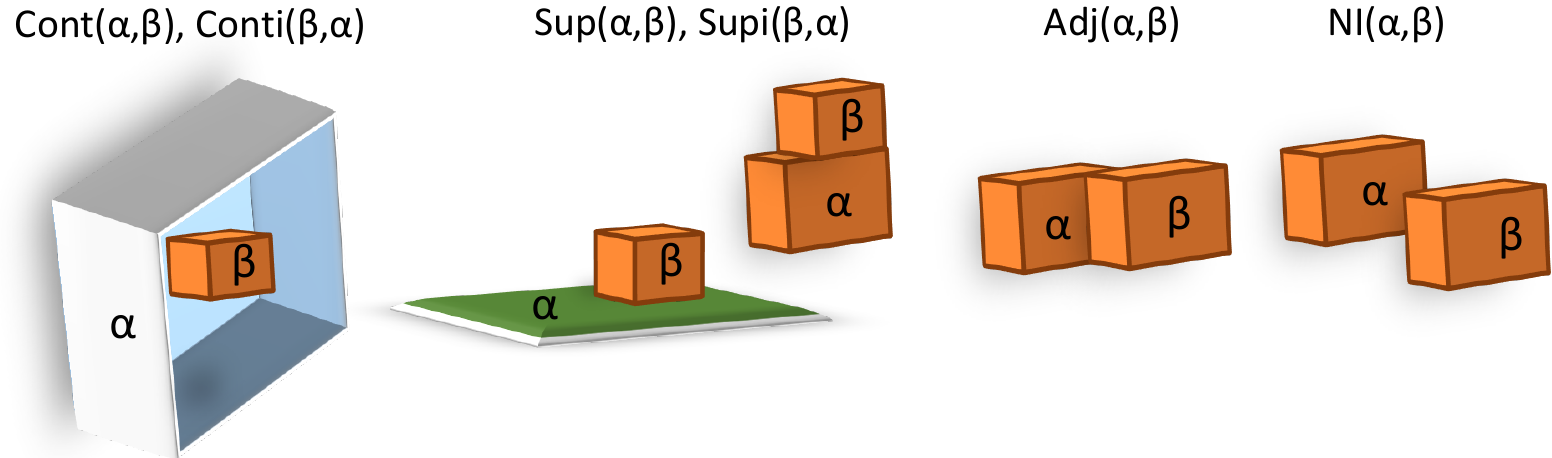}
    \caption{DiSR representations in 3D space.}
    \label{fig:disr_relations}
\end{figure}

Effectively, detecting simple and discrete spatial interactions \eg `touching', `not touching' as well as more complex ones  \eg `supporting', `containing', in the 2D image plane, is challenging, especially when there is object occlusion.
To address this limitation we propose a depth-informed set of qualitative spatial relationships, which enable the differentiation between objects' visual occlusion and actual interaction, and take into account the object's convexity-type.
The object's convexity-type carries information about the object's affordance, hence the kind of interaction which can hold between that object and another. For example, the relation \emph{contains} can hold between a concave object, due to its concavity, and another object when interacting in a specific way.
However in an interaction between objects without any visual concave curve, such a relation will not hold.

The spatial information employed to describe the object interactions in the spatial domain consists of a discrete set of spatial relations that can describe any spatial configuration of the objects in a scene.
These depth-informed interactions are exploited for the construction of more accurate qualitative graphical structures, achieving more homogeneous and complete affordance clusters.

Previous work on affordances using QSRs \cite{sridhar2008learning} has used the purely mereotopological RCC set of relations \cite{randell1992spatial}, \cite{cohn1997qualitative}. Here we propose a more expressive set of relations (DiSR), which exploit knowledge that can be garnered from a 2.5D representation: objects' convexity type and their orientation in the vertical plane.
We propose the set of \emph{Depth-informed Spatial Relations} ({\em DiSR}) (Fig. \ref{fig:disr_relations})\footnote{The inverse of a relation is denoted with an \textsf{i} at the end, wherever an inverse exists, \eg the inverse of \textsf{Sup} is \textsf{Supi}.}: `supports' (\textsf{Sup}, \textsf{Supi}), `contains' (\textsf{Cont}, \textsf{Conti}), `adjacent' (\textsf{Adj}), `not interacting'(\textsf{NI}).

Depth information is considered for reasoning about the spatial relative positions of objects in 3D space, hence enabling the distinction between occlusions and interactions.
Furthermore, from depth cues we can effectively infer the convexity-type of the detected objects, and along with the corresponding depth differences, DiSR relations are extracted.
For a spatial interaction to hold, an overlap on the depth dimension must be evident as well as the \textit{x}, \textit{y} dimension, and depending on the convexity-type of the interacting objects and in which part of the depth information the overlap appears, a different qualitative spatial relation holds from the set of DiSR.

DiSR spatial relations take into account the depth information of the detected object masks in the scene as well as their 2D location considering the bounding box enclosing the detected mask of every object.
We present in Table \ref{table:disr} the definition of every DiSR relation inspired by the RCC set, whilst exploiting the RCC relations: `overlap' (\textsf{O}) and `part' (\textsf{P}, \textsf{Pi}) computed in the 2D image plane.

\begin{table}
\renewcommand{\arraystretch}{1.2}
\centering
\begin{threeparttable}
\caption{DiSR} \label{table:disr}
\begin{tabularx}{\linewidth}{VBS}
    \rowcolor{Gray}
    \textbf{DiSR} & \textbf{Definition} & \textbf{Description}\\
    \hline
    \textsf{Sup}($\alpha$,$\beta$) & $(\textsf{DPO}(\alpha,\beta) \land \text{Surface}(\alpha)) \lor \textsf{On}(\beta,\alpha)$ & $\alpha$ supports $\beta$ \\
    \rowcolor{Gray}
    \textsf{Cont}($\alpha$,$\beta$) & $\textsf{P}(\beta,\alpha) \land \text{Concave}(\alpha) \land \textsf{DPP}(\beta, \alpha)$ & $\alpha$ contains $\beta$\\
    \textsf{Adj}($\alpha$,$\beta$) & $\textsf{O}(\alpha,\beta) \land \textsf{DPO}(\alpha,\beta) \land \neg \textsf{Cont}(\alpha,\beta) \land \neg \textsf{Cont}(\beta,\alpha) \land \neg \textsf{Sup}(\alpha,\beta) \land \neg \textsf{Sup}(\beta,\alpha)$ & $\alpha$ is adjacent to $\beta$\\
    \rowcolor{Gray}
    \textsf{NI}($\alpha$,$\beta$) & $\neg \textsf{Sup}(\alpha,\beta) \land \neg \textsf{Sup}(\beta,\alpha) \land \neg \textsf{Cont}(\alpha,\beta) \land \neg \textsf{Cont}(\beta,\alpha) \land \neg \textsf{Adj}(\alpha,\beta) \land \neg \textsf{Adj}(\beta,\alpha)$ & $\alpha$ does not interact with $\beta$\\
    \hline
    \hline
    \multicolumn{3}{c}{\textbf{\textsf{P} and \textsf{O} of RCC relations}}\\
    \rowcolor{Gray}
    \textsf{P}($\alpha$,$\beta$) & $\forall z (\textsf{C}(z,\alpha) \rightarrow \textsf{C}(z,\beta))$\tnote{\textdagger} & $\alpha$ is a part of $\beta$\\
    \textsf{O}($\alpha$,$\beta$) & $\exists z (\textsf{P}(z,\alpha) \land \textsf{P}(z,\beta))$ & $\alpha$ overlaps $\beta$\\
\end{tabularx}
\begin{tablenotes}
\item[\textdagger] \textsf{C} relation is defined in Table \ref{table:rcc2}.
\end{tablenotes}
\end{threeparttable}
\end{table}

\begin{figure}
    \centering
    \includegraphics[trim ={40mm 128mm 150mm 17mm},clip, width=1.0\textwidth]{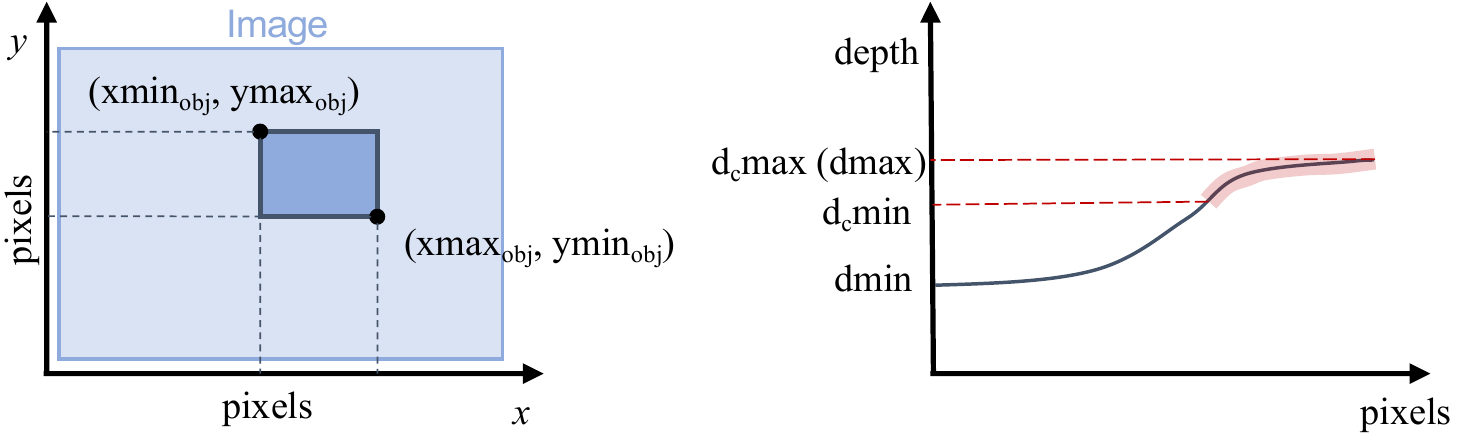}
    \caption{(left) Detected object bounding box coordinates. (right) Object depth distribution and the depth information signifying potential concave curve detection between the values \minconcave and \maxconcave (red).}
    \label{fig:bboxes}
\end{figure}

The convexity type of an object, being either \emph{concave}, \emph{surface} or \emph{convex}, is described further in \S \ref{sec:object_convexity_type_detection}. `Depth Overlap' (\textsf{DPO}) and `Depth Proper Part' (\textsf{DPP}, \textsf{DPPi}) are primitive relations which hold between the objects' depth information, defined as:
\begin{gather*}
    \textsf{DPO}(\alpha,\beta) \equiv 
    ((\textit{dmax}_\alpha \geq \textit{dmin}_\beta) \land(\textit{dmax}_\alpha < \textit{dmax}_\beta) \land  \\
    (\textit{dmin}_\alpha < \textit{dmin}_\beta)) \lor ((\textit{dmax}_\beta \geq \textit{dmin}_\alpha) \land \\
    (\textit{dmax}_\beta < \textit{dmax}_\alpha) \land (\textit{dmin}_\beta < \textit{dmin}_\alpha)) \\
    \textsf{DPP}(\alpha,\beta) \equiv (\textit{dmax}_\alpha > d_c\textit{min}_\beta) \land (\textit{dmax}_\alpha \leq d_c\textit{max}_\beta) \land \\
    (\textit{dmin}_\alpha \geq d_c\textit{min}_\beta) \land (\textit{dmin}_\alpha < d_c\textit{max}_\beta)
\end{gather*}

\noindent where \maximumdepth and \minimumdepth are the maximum and minimum depth values, respectively, by considering the depth cues of the detected object's mask, and \maxconcave and \minconcave are the maximum and minimum depth values of the detected concave curve area. Figure \ref{fig:bboxes} (right) illustrates a concave object's concave curve area ($[$\minconcave,\maxconcave$]$), where ascending-ordered depth-pixel values are on the \textit{x}-axis and their depth values on the \textit{y}-axis. The values of \minconcave and \maxconcave are obtained by Alg. \ref{alg:convexity_depth}, which is further explained in \S \ref{sec:object_convexity_type_detection}.

\noindent Moreover, the primitive spatial relation \textsf{On} is defined as:
\begin{gather*}
    \textsf{On}(\alpha,\beta) \equiv \textsf{O}(\alpha,\beta) \land ((\textit{ymax}_\alpha \geq \textit{ymax}_\beta) \land \\
    (\textit{ymin}_\alpha \geq \textit{ymin}_\beta) \land (\textit{xmax}_\alpha \leq \textit{xmax}_\beta) \land (\textit{xmin}_\alpha \geq \textit{xmin}_\beta))
\end{gather*}

\noindent where $(\textit{xmin}, \textit{ymax})$ and $(\textit{xmax}, \textit{ymin})$ are the top-left and bottom-right corners of the detected object's bounding box (Fig. \ref{fig:bboxes} (left)). 

The definitions proposed are an approximation to the meaning of the English descriptions in Table \ref{table:disr}; it is possible to satisfy the definitions with configurations which do not accord with intuitive meaning of the English word.
\Eg in the case where three objects are stacked the one on top of the other, then in the 2D image plane \textsf{On}({\em top object}, {\em bottom object}) will be \textsf{True} even though there is a {\em middle object} in between.
However, these definitions are easy to compute and work well in the everyday scenes we have considered so far. In future work we may refine these relationships to more closely correspond to the semantics of the English words.

\subsection{Object Convexity Type Detection} \label{sec:object_convexity_type_detection}

\begin{figure}
    \centering
    \subfigure{
    \includegraphics[trim ={1mm 1mm 1mm 1mm},clip, width=.18\textwidth]{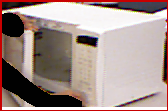}
    \put(-80,65){\textsf{Detected object}}
    \put(17,65){\textsf{Depth regions}}
    \put(100,65){\textsf{Deep depth regions}}
    \put(200,65){\textsf{Deep depth contours}}
    }
    \subfigure{
    \includegraphics[trim ={0mm 0mm 0mm 0mm},clip, width=.18\textwidth]{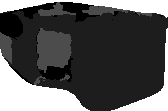}
    }
    \subfigure{
    \includegraphics[trim ={0mm 0mm 0mm 0mm},clip, width=.18\textwidth]{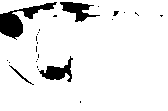}
    }
    \subfigure{
    \includegraphics[trim ={0mm 0mm 0mm 0mm},clip, width=.18\textwidth]{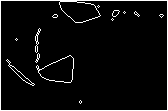}
    }
    
    \subfigure{
    \includegraphics[trim ={1mm 1mm 1mm 1mm},clip, width=.18\textwidth]{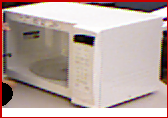}
    }
    \subfigure{
    \includegraphics[trim ={0mm 0mm 0mm 0mm},clip, width=.18\textwidth]{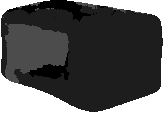}
    }
    \subfigure{
    \includegraphics[trim ={0mm 0mm 0mm 0mm},clip, width=.18\textwidth]{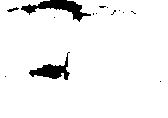}
    }
    \subfigure{
    \includegraphics[trim ={0mm 0mm 0mm 0mm},clip, width=.18\textwidth]{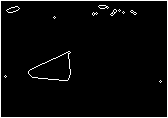}
    }
    
    \subfigure{
    \includegraphics[trim ={1mm 1mm 1mm 1mm},clip, width=.18\textwidth]{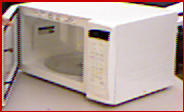}
    }
    \subfigure{
    \includegraphics[trim ={0mm 0mm 0mm 0mm},clip, width=.18\textwidth]{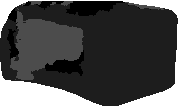}
    }
    \subfigure{
    \includegraphics[trim ={0mm 0mm 0mm 0mm},clip, width=.18\textwidth]{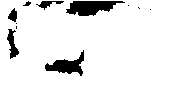}
    }
    \subfigure{
    \includegraphics[trim ={0mm 0mm 0mm 0mm},clip, width=.18\textwidth]{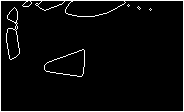}
    }
    
    \subfigure{
    \includegraphics[trim ={1mm 1mm 1mm 1mm},clip, width=.18\textwidth]{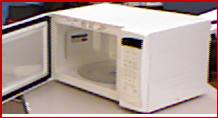}
    }
    \subfigure{
    \includegraphics[trim ={0mm 0mm 0mm 0mm},clip, width=.18\textwidth]{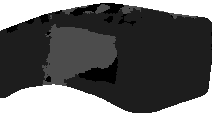}
    }
    \subfigure{
    \includegraphics[trim ={0mm 0mm 0mm 0mm},clip, width=.18\textwidth]{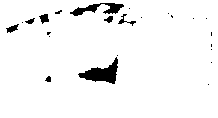}
    }
    \subfigure{
    \includegraphics[trim ={0mm 0mm 0mm 0mm},clip, width=.18\textwidth]{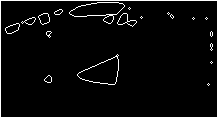}
    }
    
    \subfigure{
    \includegraphics[trim ={1mm 1mm 1mm 1mm},clip, width=.18\textwidth]{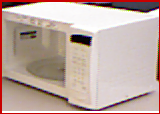}
    }
    \subfigure{
    \includegraphics[trim ={0mm 0mm 0mm 0mm},clip, width=.18\textwidth]{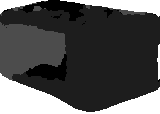}
    }
    \subfigure{
    \includegraphics[trim ={0mm 0mm 0mm 0mm},clip, width=.18\textwidth]{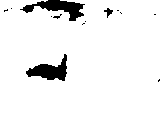}
    }
    \subfigure{
    \includegraphics[trim ={0mm 0mm 0mm 0mm},clip, width=.18\textwidth]{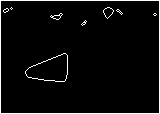}
    }
    
    \subfigure{
    \includegraphics[trim ={1mm 1mm 1mm 1mm},clip, width=.18\textwidth]{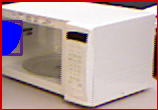}
    }
    \subfigure{
    \includegraphics[trim ={0mm 0mm 0mm 0mm},clip, width=.18\textwidth]{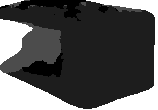}
    }
    \subfigure{
    \includegraphics[trim ={0mm 0mm 0mm 0mm},clip, width=.18\textwidth]{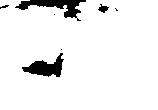}
    }
    \subfigure{
    \includegraphics[trim ={0mm 0mm 0mm 0mm},clip, width=.18\textwidth]{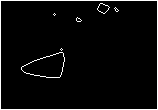}
    }
    
    \subfigure{
    \includegraphics[trim ={1mm 1mm 1mm 1mm},clip, width=.18\textwidth]{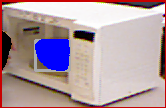}
    }
    \subfigure{
    \includegraphics[trim ={0mm 0mm 0mm 0mm},clip, width=.18\textwidth]{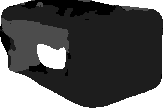}
    }
    \subfigure{
    \includegraphics[trim ={0mm 0mm 0mm 0mm},clip, width=.18\textwidth]{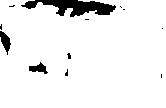}
    }
    \subfigure{
    \includegraphics[trim ={0mm 0mm 0mm 0mm},clip, width=.18\textwidth]{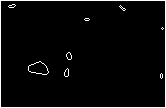}
    }
    \caption{Processing steps for extracting depth contours. The bounding boxes of the detected objects in the RGB frames are mapped onto the depth data from where we extract all the depth information partitioned into sections depending on their depth values. Finally, from that depth data, deep regions are considered from which depth contours are created indicating potential \emph{indentation areas}, thus this object is classified as `concave'. The extraction of deep depth regions is based on the employment of the \depththreshold threshold. Any information indicating another detected object occluding the primary one, is discarded; the last two rows illustrate the presence of a bowl object (blue) whose depth information is not considered as shown in the `Depth regions' column.}
    \label{fig:object_concavity}
\end{figure}

\begin{algorithm}[t]
	\caption{Compute the convexity type of an object.}\label{alg:object_convexity}
	\hspace*{\algorithmicindent} \textbf{Given:} \depththreshold
	\begin{algorithmic}[1]
		\Procedure{ObjectConvexity}{$\text{dist}_{\text{depth}}$}
		\State \maximumdepth $\gets max(\textit{dist}_{\textit{depth}})$; \minimumdepth $\gets min(\textit{dist}_{\textit{depth}})$
		\State $C \gets \text{ContourHierarchy}(\textit{dist}_{\textit{depth}})$
		\If {$\bigl( ($ \maximumdepth $-$ \minimumdepth $) < $ \depththreshold $\bigl)$ \& $\bigl( C.child()$ exists $\bigl)$}
		\State $\textit{object}_{\textit{type}} \gets \text{concave}$
		\Else
		
		\If{$($ \maximumdepth $-$ \minimumdepth $) < $ \depththreshold}
		\State $\textit{object}_{\textit{type}} \gets \text{surface}$
		\Else
		\ $\textit{object}_{\textit{type}} \gets \text{convex}$
		\EndIf
		\EndIf
		\State \textbf{return} $\textit{object}_{\textit{type}}$
		\State \textbf{end}
		\EndProcedure
	\end{algorithmic}
\end{algorithm}

By considering the depth information we obtain knowledge about the \textit{indentation area} ({\textsf{m-}}) and \textit{protrusion area} ({\textsf{M+}}) of an object, as defined in the \textit{Process-Grammar} \cite{leyton1988process}.
More specifically, visually `convex' type objects do not appear to have any \textit{indentation areas} whereas `concave' type objects are characterized by their concavity curve, which morphologically appears as a bay-formation described as {\textsf{M+}}{\textsf{m-}}{\textsf{M+}}.
Such information is critical to ascertain a relation when two objects interact in the 2D image plane, \eg a {\textsf{Cont}} DiSR occurs between one or more `concave' type objects when the depth information of the \emph{containee} confirms that is between {\textsf{m-}} and {\textsf{M+}} areas of the \emph{container}.
In this section we will demonstrate how we derive the convexity type of the detected objects from the depth cues.

\begin{figure}
    \centering
    \includegraphics[trim ={9mm 105mm 76mm 8mm},clip, width=1.\textwidth]{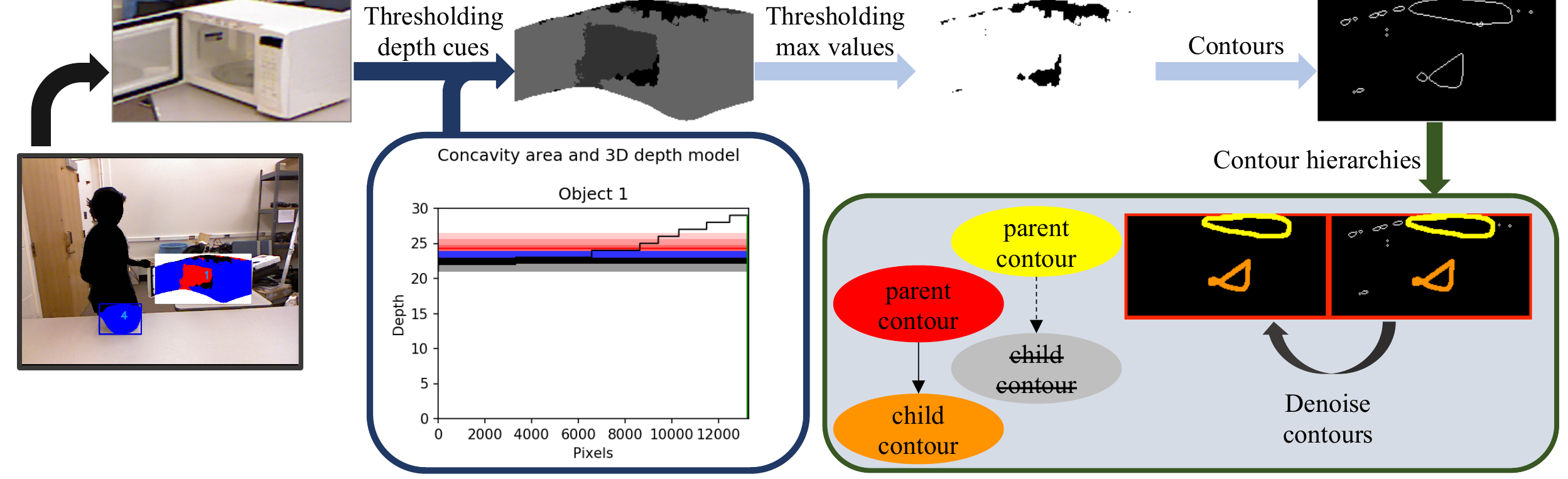}
    \caption{Process steps for detecting `concave' type objects. From an RGB video frame objects are detected using an object detector; numbers are shown in the image frame for every detected object. The bounding boxes proposing an object detection are mapped on the depth domain to extract the depth data relevant to that object detection. The depth data is then partitioned and deep depth regions are exploited to produce depth contour's of possible concave curve areas. Finally, depth contours are a hierarchical set depending on each contour's inclusion, \eg the orange contour is included in the red contour hence it is a child of the red contour. If a child contour exists then the region it encapsulates is detected as a concave curve area.}
    \label{fig:concave_detection}
\end{figure}

We identify three primary convexity type objects: `concave', `convex', and `surface' by employing Alg. \ref{alg:object_convexity}.
For the determination of the objects' convexity type the increasing-ordered-by-value depth information across the pixels defining each object is considered (Fig. \ref{fig:bboxes}(right)).
Alg. \ref{alg:object_convexity} is based on a convexity depth threshold (\depththreshold), which defines the upper boundary of depth range information of a `convex' type object. Objects with depth range (considering the minimum and maximum values of the $\text{dist}_{\text{depth}}$) greater than \depththreshold are subject to be grouped under `concave' or `surface' depending on their depth contour hierarchies (\textit{ContourHierarchy}).
A depth contour is the contour created by the depth information exceeding \depththreshold (Fig. \ref{fig:object_concavity}).
Contour hierarchies establish a tree structure of contour inclusion, where every node of the tree stands for a contour and every parent includes its children.
In this case, the root of the tree is always the frame of the image. Contours which are not directly included in the root contour are not considered its children.
We eliminate potential contour noise by considering the relative difference of the size of the contours with respect to the detected object. The remaining contours are then examined in terms of hierarchy. A \textit{child contour} indicates an \textit{indentation area}.
Thus, the detection of a child contour in the depth domain, indicates the presence of a concave curve, therefore a `concave' type object, and `surface' type otherwise\footnote{This approach considers this restricted representation for detecting concave objects as the depth resolution varies depending on the objects' distance from the visual sensor.}.
Figure \ref{fig:concave_detection} summarizes the process of the detection of a concave object.

\begin{algorithm}[t]
	\caption{Define min and max depth of an object's concavity.} \label{alg:convexity_depth}
	\hspace*{\algorithmicindent} \textbf{Given:} $h$, $n$
	\begin{algorithmic}[1]
		\Procedure{ConvexityDepth}{$\text{dist}_{\text{depth}}$}
		\State \maximumdepth $\gets max(\textit{dist}_{\textit{depth}})$; 	 \minimumdepth $\gets min(\textit{dist}_{\textit{depth}})$
		\State $\textit{object}_{\textit{type}} \gets \text{\small{O}\scriptsize{BJECT}\small{C}\scriptsize{ONVEXITY}}(\textit{dist}_{\textit{depth}})$
		\If{$\textit{object}_{\textit{type}} = \text{concave}$}
		\State $\textit{sections}  \gets ($ \maximumdepth $-$ \minimumdepth $)/h$
		\State \maxconcave $\gets$ \maximumdepth
		\State \minconcave $\gets$ \maximumdepth $- (n *$ \textit{sections}$)$
		\Else \ 
		\maxconcave $\gets$ \maximumdepth; 
	\minconcave $\gets$ \minimumdepth
		\EndIf
		\State \textbf{return} \maxconcave, \minconcave
		\State \textbf{end}
		\EndProcedure
	\end{algorithmic}
\end{algorithm}

However, defining only the type of an object is not sufficient for reasoning about the spatial interactions between objects. Detecting complex spatial object relations effectively, \ie contains, is challenging without considering potential concave curves of objects.
Thus, we propose Alg. \ref{alg:convexity_depth} to infer the boundaries of the {\textsf{m-}} and {\textsf{M+}} areas, as a direct generalization of the \emph{Process-Grammar} to 2.5D, indicating a concave curve in 3D space, with respect to the object's depth information and convexity type.
For a `concave' type object we partition the ordered-by-value depth information into $h$ sections for distinguishing the \textit{indentation} from the \textit{protrusion area}.
The $n$ sections with the highest depth values are estimated to capture its concave curve.
We set these depth boundaries of such objects to enclose the potential concave curve's depth information for detecting the relation {\textsf{Cont}}.
The parameterization of $h$ $n$, and \depththreshold was conducted in an empirical study and is further explained in \S \ref{sec:experimental_setup}.
Depth boundaries of `convex' and `surface' type objects are not processed due to concave curve absence. \maxconcave and \minconcave set the boundaries of the potential concave curve area of a `concave' type object and are further employed for detecting a \emph{contains} relation between a pair of objects.


\section{Learning Object Affordances} \label{sec:learning_object_affordances}

We learn to categorize object affordances by clustering \emph{AGraphlets} whilst employing a hierarchical approach. Every \emph{AGraphlet} represents the interaction between a pair of objects in the scene. We consider an interaction between two objects as the spatio-temporal sequence of relations holding during an activity.
By clustering such graph structures we produce a hierarchy of similar affordances, which are closely related with the way every object is being used in an activity. Hence, our method does not pose any constraints on the number of affordance clusters an object can be assigned to, whilst every object has as many \emph{AGraphlets} as detected interactions.
\Eg consider the scenario where an agent picks a bowl from a table and places it in the microwave.
The clustering mechanism considers the interaction of the bowl with the table and the bowl with the microwave as two different interactions; two \emph{AGraphlets} will be created for the bowl implying that it is a \emph{supportable} as well as a \emph{containable} object.

As \emph{AGraphlets} are high-level object interaction representations defined in the space of the qualitative relations being captured, computing the similarity between them is based on the spatio-temporal relations occurring. To effectively cluster these representations we project the \emph{AGraphlets} in a \textit{d}-dimensional latent space and perform clustering on that space instead. The \emph{graph2vec} network is employed for this purpose, creating \textit{d}-dimensional graph embeddings for each \emph{AGraphlet}.

To compute the similarity/dissimilarity of these graph embeddings we exploit a cost function using the cosine similarity measure, defined as:

\begin{equation}
C(\mathrm{A}, \mathrm{B}) = 1- cos(\mathrm{A}, \mathrm{B}) = 1 - \frac{\mathrm{A} \cdot \mathrm{B}}{||\mathrm{A}|| \times ||\mathrm{B}||}
\label{eq:cosine}
\end{equation}

\noindent where the values of the $cos()$ function range from -1 to 1, where 1 indicates that the vector embeddings are exactly the same, 0 that they are orthogonal, and -1 means that the vectors are exactly opposite.
In the \emph{AGraphlet} space, a 1 cosine value stipulates that the two graphs carry the same spatio-temporal information, hence the same affordance, whereas a value of -1 suggests that the two graphs have no spatio-temporal information in common, thus each one of them represents a completely different affordance. A cosine value of 0 denotes that the two graphs are opposite, each one carrying information of the two objects that interact.

We perform a hierarchical clustering on the graph embeddings, using the cost function defined in Eq.~\ref{eq:cosine}, to produce a dendrogram of similarities of the extracted \emph{AGraphlets}. Clusters are formed by grouping the leaves of the dendrogram with respect to their hierarchy. Figure \ref{fig:object_hierarchies}(top) illustrates a subset dendrogram of the complete hierarchical clustering output.
Such a hierarchy reveals the similarities of the different interactions occurring in the data, which thus yields a set of affordance clusters. Different set of clusters are formed based on a tunable threshold in our hierarchy. The higher in the hierarchy we consider creating the clusters, the more general the clusters become in terms of affordances.


\section{Leeds Object Affordance Dataset (\emph{LOAD})} \label{sec:leeds_object_affordance_dataset_load}

\begin{figure}
    \centering
    \includegraphics[trim ={0mm 137mm 0mm 0mm},clip, width=.9\linewidth]{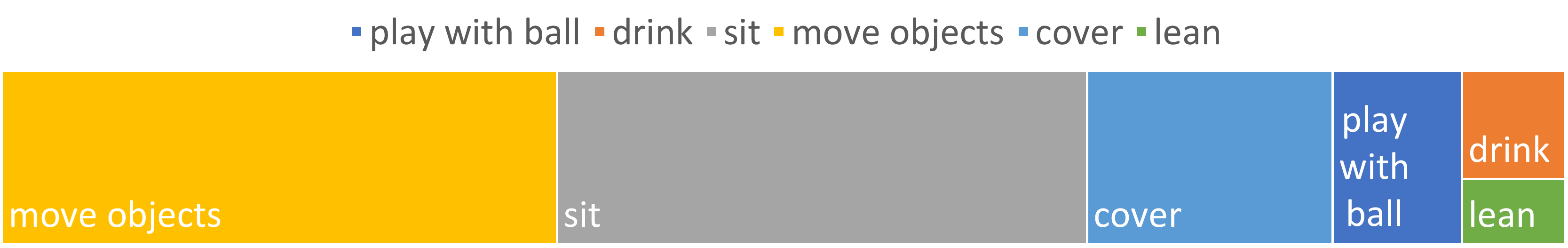}
    \caption{Coarser grouping of activities in LOAD dataset.}
    \label{fig:load_coarse_activities}
\end{figure}

\begin{figure}
    \centering
    \includegraphics[trim ={1mm 79mm 14mm 1mm},clip, width=1.\linewidth]{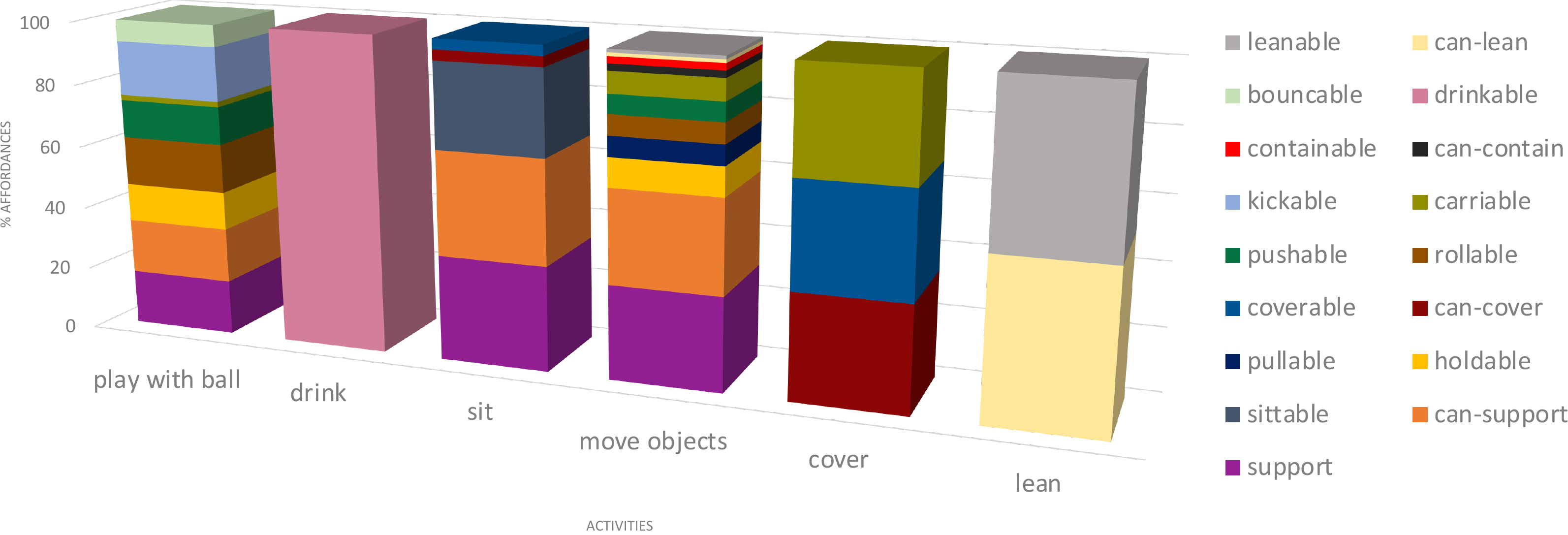}
    \caption{Analysis of the percentage of affordances for every activity, considering a coarser grouping of activities.}
    \label{fig:load_affordances_activities}
\end{figure}

We present the Leeds Object Affordance Dataset or LOAD dataset, which effectively extends the domain of activities existing in all the publicly available datasets to date.
This newly presented dataset comprises of 58 RGB-D videos with an average number of 258 frames per video, capturing various human-object interactions, which are not present in the existing datasets visualizing human-object interactions. LOAD dataset aims to push the boundaries of affordance distinction and categorization through more complex real-world activities.
Sample frames presenting the different object affordances captured can be found in the appendix.
Figure \ref{fig:load} illustrates the percentage of the affordance labels occurring in the LOAD dataset and Figure \ref{fig:load_coarse_activities} presents the percentage of different kinds of activities. Moreover, Figure \ref{fig:load_affordances_activities} displays the percentage of occurrence of the affordances in every kind of activity. The main benefit of the LOAD dataset, as shown in these figures, is that the same affordance is present in different kind of activities and between different objects. From Figure \ref{fig:load_affordances_activities}, though some affordances are present in only a single activity, \ie `drinkable', there are others that are spread throughout different kind of activities, \ie `support', `can-support', `holdable', `rollable', `coverable', `can-cover'.
In this dataset, apart from the interactive objects, we also consider the `floor' as an entity that holds an affordance, thus the affordance labels of `can-support' and `supportable' dominate in the ground truth. Also, there are several affordance labels, which do not have a complementary affordance. This is due to affordances related to human-object interactions, \eg `kickable', `pushable', `pullable', etc.

\begin{figure}
    \centering
    \includegraphics[trim ={5mm 15mm 5mm 18mm},clip, width=1.\linewidth]{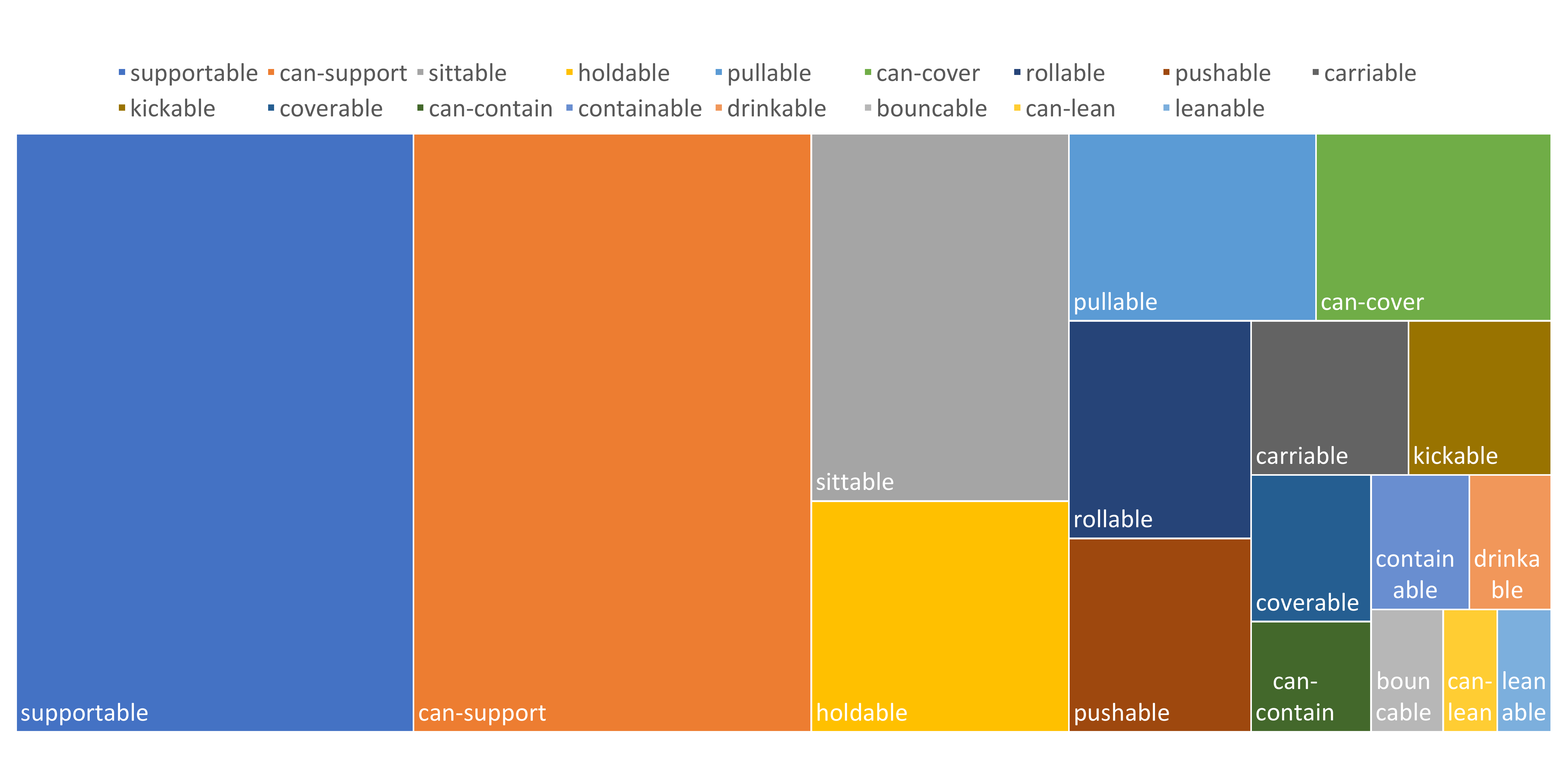}
    \caption{Affordance labels in the LOAD dataset; area of rectangle reflects abundance in the dataset.}
    \label{fig:load}
\end{figure}

\section{Experimental Setup} \label{sec:experimental_setup}

\subsection{Datasets} \label{sec:datasets}

For the evaluation of our method we found the publicly available CAD-120 dataset \cite{koppula2013learning} (CAD) suitable for the task of object affordance categorization. Moreover, to prove generalizability across different data distributions we also tested our method on the Watch-n-Patch \cite{wu2015watch} (WnP), and on the LOAD dataset (LOAD) newly presented here.
All three datasets comprise activities of everyday-life scenarios with various configurations of the objects in the scene as well as different camera orientations. Table \ref{table:datasets} summarizes the characteristics of the datasets employed in this study.
Moreover, the affordance labels present in the LOAD dataset are indicated in Figure \ref{fig:load}, whereas the labels for both the CAD-120 and Watch-n-Patch datasets are: `can-contain', `containable', `can-support', `supportable', `drinkable', and `holdable'.

We used 80\% of the CAD-120 dataset to determine the method's parameters and train the \emph{graph2vec} network, resulting to 18,072 \emph{AGraphlets}, and then evaluated the proposed approach on the remaining 20\% unseen videos (24 videos), which comprise of 5,682 \emph{AGraphlets}.
Experiments on the Watch-n-Patch and LOAD datasets were conducted on 24 and 15 hand-picked videos, or 6,900 and 1,212 \emph{AGraphlets} respectively, where the predictions of the object detector, after visual inspection, were sufficient for capturing interacting objects. For defining the parameters of the algorithm, we exploited hand-picked videos from the training set for the CAD-120 dataset, and for the LOAD and Watch-n-Patch datasets we hand-picked videos different from the ones used for testing the proposed method.

\setlength{\tabcolsep}{4pt}
\begin{table}[h]
\renewcommand{\arraystretch}{1.2}
\centering
\begin{threeparttable}
\caption{Datasets} \label{table:datasets}
\begin{tabularx}{\linewidth}{a|cccccn}
    \rowcolor{Gray}
    Dataset & data & \#vid. & avg.fr.\tnote{a} & a.p.v.\tnote{b} & a.r.\tnote{c} & activities\\
    \hline\hline\noalign{\smallskip}
    CAD-120 & 2.5D & 120 & 525 & 1-2 & 3 & `arranging objects', `cleaning objects', `having meal', `making cereal', `microwaving food', `picking objects', `stacking objects', `taking food', `taking medicine', `unstacking objects' \\
    \noalign{\smallskip}
    \noalign{\smallskip}
    Watch-n-Patch & 2.5D & 457 & 170 & 1-2 & 3 & `turn on monitor', `turn off monitor', `walking', `play computer', `reading', `fetch book', `put back book', `take item', `put down item', `leave office', `fetch from fridge', `put back to fridge', `prepare food', `microwaving', `fetch from oven', `pouring', `drinking', `leave kitchen', `move kettle', `fill kettle', `plug in kettle'\\
    \noalign{\smallskip}
    \noalign{\smallskip}
    LOAD & 2.5D & 58 & 258 & 3-4 & 3 & `pull chair', `push chair', `sit on chair', `sit on the floor', `sit on the table', `push table', `pull table', `hold bottle', `hold umbrella', `carry bag', `drinking', `put item down', `kick ball', `roll ball', `bounce ball', `throw ball', `cover', `put into', `take out', `lean against the wall', `roll bottle', `roll chair', `roll suitcase', `carry chair'\\
\end{tabularx}
\begin{tablenotes}
\item[a] average number of frames (per video)
\item[b] activities per video
\item[c] activity repetition
\end{tablenotes}
\end{threeparttable}
\end{table}
\setlength{\tabcolsep}{1.4pt}

\subsection{Human/Object Detection \& Tracking} \label{sec:human_object_detection_and_tracking}

To obtain human skeletal data in the scene we compared the output of the Convolutional Pose Moachine (CPM) \cite{wei2016convolutional} model with the kinect skeletal data. Through an empirical study we found that the CPM predictions are more accurate, especially when human-joint occlusion occurs or the human agent stands on their side.

Object locations and depth information are provided from the predicted objects' masks. We employ the state-of-the-art Mask R-CNN framework \cite{he2017mask} trained on the COCO dataset \cite{lin2014microsoft}, since the training data distribution is similar to the data distribution present in the datasets we use in this work, and it is a generic dataset with more object classes than those included in the target task.
The box enclosing the object's mask corresponds to the object's bounding box.
This implementation is based on Mask R-CNN predictions, however the proposed method is not object-specific and any class-agnostic proposal method can be used.

Whilst Mask R-CNN objects' bounding boxes predictions are temporally sparse, we enrich the objects tracks by considering the CSRT tracker's predictions \cite{bolme2010visual} from the latest object bounding box occurrence, for frames where Mask R-CNN failed to detect the object, hence creating tracks of class agnostic-masks.
CSRT tracker solely relies its predictions of the visual characteristics of the initial object mask predictions, considering only past instances on mask detections. Hence, it is able to track non-rectangular objects, which aids in having more accurate tracking predictions by considering the objects mask's rather than their bounding boxes.
A threshold of minimum $0.5$ IoU overlap, determined from an empirical study, is required to assign a bounding box prediction as the predicted location of a detected object.

The predicted object mask's depth information is employed to infer the object's convexity type. We exclude any pixel, which is part of a human detection, by retrieving a human mask from the DensePose framework \cite{alp2018densepose}.
Whilst Mask R-CNN produces object masks for every object separately, overlapping objects have overlapping masks, thus the intersected mask area may cause problems in determining the object's convexity type. A \emph{semantic depth map} is constructed for every frame of the video data, consisting of all the predicted object masks while eliminating any detected intersection mask area. This is achieved by assigning every pixel of such area to the object with the highest mask detection score.

\subsection{Parameters} \label{sec:parameters}

We employed the QSRlib library \cite{gatsoulis2016qsrlib} for the construction of \emph{AGraphlets}.
For the $h$ and $n$ values which are used in defining the \textit{indentation area} of a concave object, we selected the values of 5 and 3 respectively, after conducting an empirical study for $h \in \{2,3,4,5,6,7\}$ and $n \in \{1,\dots, h-1\}$. We also set the \depththreshold value to be 4 for the CAD-120 dataset after evaluating on the values 1,2,3,4,5,6,7,8,9,10, 0.3 for the Watch-n-Patch after consideration of the values in the range 0.1 to 2.0 with step 0.1, and 10.0 for the LOAD after studying the values in the range 8 to 20 with step 1.

For the creation of clusters we threshold the produced hierarchy at 0.02 for all three datasets. The value of 0.03 was employed for the RCC5(+ON) experiment (\S \ref{sec:quantitative_evaluation}), though for the sED (\S \ref{sec:quantitative_evaluation}) the value 1.0 was used. This hierarchy threshold was defined by evaluating the Bayesian Information Criterion (BIC) and Akaike Information Criterion (AIC) on the training set for different heights of the hierarchy.

\subsection{\emph{graph2vec} Network} \label{sec:graph2vec_network} 

We trained from \emph{graph2vec} network with the \emph{AGraphlets} of the training set to create a 128-dimensional latent space where the interaction graphs can be projected on. We employed the network architecture provided by \cite{narayanan2017graph2vec} and we trained it using Stochastic Gradient Descent, a batch size of 512, and setting the learning rate to 0.5. Also, we set the WL subgraph extraction value to 14 which, through an empirical study, we found it was sufficient enough to create all the different subgraphs from the \emph{AGraphlets}.

The selection of the embedding size and learning rate was made from two empirical studies detailed in the appendix.


\section{Experimental Evaluation}

\subsection{Quantitative Evaluation} \label{sec:quantitative_evaluation}

In this section, we will analyze the experimental results by inspecting the clusters of affordances formed, and evaluating their homogeneity and completeness.
The datapoints clustered consist of detected objects with their interactions with any other object and the human agent, allowing multiple datapoints to point to the different interactions a single object might hold. We evaluate the clusters by reporting the normalized \emph{V-measure}, \emph{homogeneity}, and \emph{completeness} scores, with higher values implying a better clustering.
Homogeneity captures how homogeneous the clusters are with respect to the groundtruth annotations, whereas completeness expresses how complete the clusters are in terms of the groundtruth affordance labels, \ie encapsulating as many as possible of the same groundtruth labels into a single cluster. The V-measure represents an average of the homogeneity and completeness scores.

\subsubsection{Baselines Experiments} \label{sec:baselines_experiments}

\setlength{\tabcolsep}{4pt}
\begin{table}
\centering
\begin{center}
\caption{Baselines experiments.}
\label{table:experiments_baselines}
\begin{tabular}{cc|c|cc}
\hline\noalign{\smallskip}
& Method & V-measure & Homogeneity & Completeness\\
\noalign{\smallskip}
\hline \hline
\noalign{\smallskip}
\multirow{3}{*}{\rotatebox{90}{CAD}} & RCC5(+ON) & 0.80 & 0.93 & 0.70 \\
                                         & sED & 0.57 & 0.67 & 0.49 \\
                                         & Ours & \textbf{0.87} & \textbf{0.99} & \textbf{0.77} \\
\noalign{\smallskip}
\hline
\noalign{\smallskip}
\multirow{3}{*}{\rotatebox{90}{WnP}} & RCC5(+ON) & 0.49 & 0.83 & 0.34 \\
                                               & sED & 0.14 & 0.13 & 0.16 \\
                                               & Ours & \textbf{0.57} & \textbf{0.99} & \textbf{0.40} \\
\noalign{\smallskip}
\hline
\noalign{\smallskip}
\multirow{3}{*}{\rotatebox{90}{LOAD}} & RCC5(+ON) & 0.62 & 0.78 & 0.52 \\
                                      & sED & 0.40 & 0.44 & 0.37 \\
                                      & Ours & \textbf{0.70} & \textbf{0.99} & \textbf{0.54} \\
\hline
\end{tabular}
\end{center}
\end{table}
\setlength{\tabcolsep}{1.4pt}

We compare our method with the baselines presented in Table \ref{table:experiments_baselines}, in which we evaluate the impact of using depth information to enhance 2D qualitative spatio-temporal relations, and projecting \emph{AGraphlets} into a latent space. The baselines are:

\begin{itemize}
    \item RCC5(+ON): we evaluated our method with the RCC5 \cite{randell1992spatial}, which consists of the relations: `discrete' (\textsf{DR}), `partially overlapping' (\textsf{PO}), `proper part' (\textsf{PP}, \textsf{PPi}), and `equal' (\textsf{EQ}), along with the addition of the \textsf{On} spatial relation defined in \S \ref{sec:formulation_of_disr}. We aim to investigate how the lack of depth information affects the affordance categorization. For this study we keep the method's configuration unchanged and conduct the experiments altering only the exploited spatial relations;
    
    \item sED: we performed hierarchical clustering on the \emph{AGraphlet} without projecting them in a latent space learned by the \emph{graph2vec} network. This baseline evaluates the clustering performance by considering a vanilla measure for graph comparison and aims to showcase the contribution of projecting \emph{AGraphlets} in a latent space. Hence we exploit a \emph{set Edit Distance} measure (Eq. \ref{eq:sed}, where $c_{spat}$ and $k_{spat}$ are 0.5 from an ablation study) to create clusters of similar affordances;
\begin{equation}
\begin{split}
sED = c_{spat}{\sum}_{v \in \eqsubgraphmultivertices[\alpha][\beta][DiSR] } {v} + c_{temp}{\sum}_{v \in \eqsubgraphmultivertices[\alpha][\beta][DiSR_{temp}] } {v} + \\
k_{spat}{\sum}_{v \in \eqsubgraphmultivertices[\alpha][\beta][RCC2] } {v} + k_{temp}{\sum}_{v \in \eqsubgraphmultivertices[\alpha][\beta][RCC2_{temp}] } {v} \\
\text{where } c_{temp} = 1 - c_{spat} \text{ and } k_{temp} = 1 - k_{spat}
\label{eq:sed}
\end{split}
\end{equation}

\begin{equation}
\begin{split}
\eqsubgraphmultivertices[\alpha][\beta][R] = \{ v : v \in \{ \eqsubgraphvertices[\alpha][R] \setminus \eqsubgraphvertices[\beta][R] \} \cup \{ \eqsubgraphvertices[\beta][R] \setminus \eqsubgraphvertices[\alpha][R] \} \}\\
\text{where } R \in \{DiSR, DiSR_{temp}, RCC2, RCC2_{temp} \}
\label{eq:vertices_difference}
\end{split}
\end{equation}
\end{itemize}

From this study, it is evident that our method outperforms the RCC5(+ON) and sED baselines in all reported metrics and datasets.
The comparison between the experimental results of the proposed approach and the RCC5(+ON) baseline, demonstrate that the employment of depth information for inferring object affordances has a considerable improvement of the V-measure in comparison to the primary RCC5 set. This improvement results from more accurate spatial relationships of interactions, enabling the creation of more accurate graphical structures for describing them. Also, the scores' improvement involving the employment of the \emph{graph2vec} network, in comparison to the sED baseline, indicates the enhancement of the clustered datapoint representation by projecting the \emph{AGraphlets} in the latent space.

\subsubsection{Related-Works Comparison} \label{sec:related-works_comaprison}

\setlength{\tabcolsep}{4pt}
\begin{table}
\centering
\begin{center}
\caption{Experimental comparison with related works.}
\label{table:experiments_relatedworks}
\begin{tabular}{cc|c|cc}
\hline\noalign{\smallskip}
& Method & V-measure & Homogeneity & Completeness\\
\noalign{\smallskip}
\hline \hline
\noalign{\smallskip}
\multirow{3}{*}{\rotatebox{90}{CAD}} & Sridhar et al. & 0.40 & 0.66 & 0.29 \\
                                         & Aksoy et al. & 0.71 & 0.84 & 0.61 \\
                                         & Ours & \textbf{0.87} & \textbf{0.99} & \textbf{0.77} \\
\noalign{\smallskip}
\hline
\noalign{\smallskip}
\multirow{3}{*}{\rotatebox{90}{WnP}} & Sridhar et al. & 0.39 & 0.65 & 0.28 \\
                                     & Aksoy et al. & \textbf{0.58} &  0.70 & \textbf{0.50} \\
                                     & Ours & 0.57 & \textbf{0.99} & 0.40 \\
\noalign{\smallskip}
\hline
\noalign{\smallskip}
\multirow{3}{*}{\rotatebox{90}{LOAD}} & Sridhar et al. & 0.63 & 0.82 & 0.51 \\
                                     & Aksoy et al. & 0.61 &  0.77 & 0.50 \\
                                      & Ours & \textbf{0.70} & \textbf{0.99} & \textbf{0.54} \\
\hline
\end{tabular}
\end{center}
\end{table}
\setlength{\tabcolsep}{1.4pt}

Related works exploit the distance between objects as a measure for perceiving their interactions, or simplistic qualitative relations, which only describe the relation of the objects with the human hand, \eg `hand approaching', `hand leaving', `in use', `idle, and `close to'.
Both approaches are not able to capture complex object interactions.
For this reason, our evaluation considers methods, which leverage qualitative spatial relations for describing human-object and object-object interactions, which are more descriptive in the spatial domain.

We compare our approach with two state-of-the-art methods that use qualitative spatio-temporal relations to describe in a high-level way human-object and object-object interactions, similar to the proposed method. These works are:

\begin{itemize}
    \item Sridhar et al.: we conducted experiments using a re-implementation of the approach of \citeauthor{sridhar2008learning} \citeyear{sridhar2008learning} which, to the best of our knowledge, is the most closely related work that exploits QSRs to capture functional object clusters;
    \item Aksoy et al.: we also performed a comparison of the proposed method with a re-implementation of the approach of \citeauthor{aksoy2010categorizing} \citeyear{aksoy2010categorizing}, which exploits high-level qualitative graphs without the employment of QSRs, to describe interactions between objects.
\end{itemize}

Both of these works employ a set of spatial qualitative relations. However, these relations are not descriptive enough to distinguish between complex object interactions,
\ie containment and support. Hence, differentiation of affordances due to different object types is not present, resulting to a coarser-grain of affordance categorization.
Moreover, occlusion is not handled, thus capturing many false positive object interactions.

Table \ref{table:experiments_relatedworks} shows the quantitative results of the experiments we conducted to demonstrate the performance of the proposed approach against the \citeauthor{sridhar2008learning} \citeyear{sridhar2008learning} and \citeauthor{aksoy2010categorizing} \citeyear{aksoy2010categorizing} baselines. Our experiments demonstrate a significant increase in all the reported metrics against the \citeauthor{sridhar2008learning} \citeyear{sridhar2008learning} work, which results from the representation of \emph{AGraphlets} that exploit more information about the interactive objects. We also achieve higher scores than the \citeauthor{aksoy2010categorizing} \citeyear{aksoy2010categorizing} work in the CAD-120 and LOAD datasets, and obtain comparable V-measure score on the Watch-n-Patch dataset. This is due to the fact that not many interactions are evident in the video scenes of the Watch-n-Patch, and the set of relations being captured are more coarse-grained, resulting to a higher completeness score, however degrading the homogeneity of the clusters.

Our results indicate that our method creates more fine-grained groups of affordances than the groundtruth annotations, \ie there can be more than one group of datapoints indicating a `supportable' affordance. Hence, we can distinguish between different objects with the same convexity type.

\subsection{Qualitative Evaluation} \label{sec:qualitative_evaluation}

\begin{figure}
    \centering
    \subfigure{
    \includegraphics[trim ={45mm 5mm 45mm 7mm},clip, width=.95\textwidth]{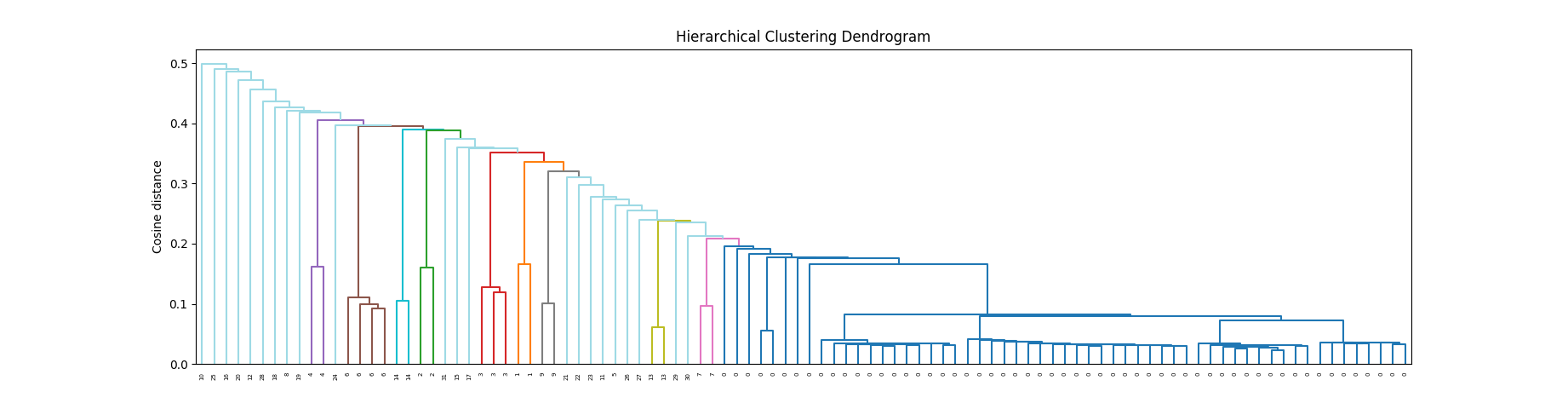}
    }
    \vfill
    \subfigure{
    \includegraphics[trim ={60mm 28mm 37mm 38mm},clip, width=.5\textwidth]{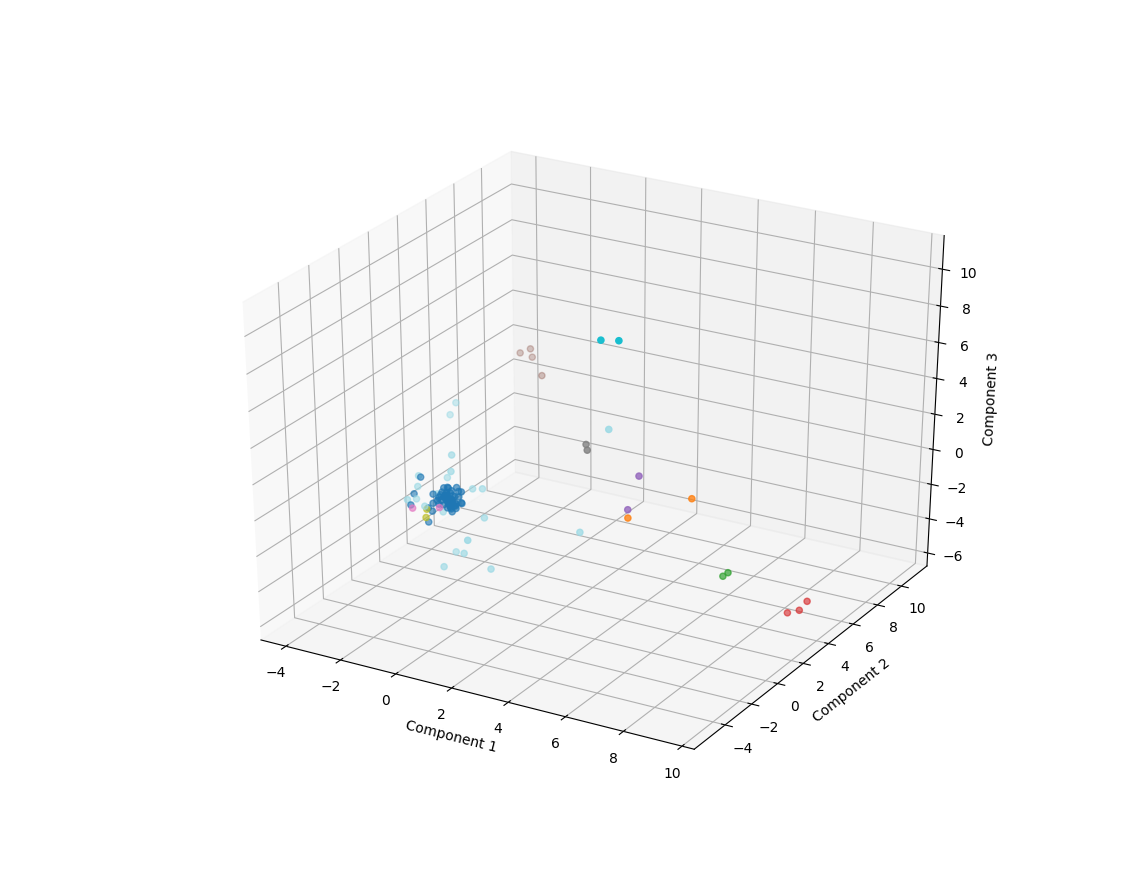}
    }
    \hfill
    \subfigure{
    \scalebox{0.7}{%
    \raisebox{5cm}{
        \begin{tabular}{||c | c||}
        \hline
        \textbf{Cluster number} & \textbf{Groundtruth label}\\
        \hline \hline
        0 & containable, supportable\\
        1 & can-contain\\
        2 & can-contain\\
        3 & unknown label\\
        4 & can-contain\\
        6 & can-support\\
        7 & supportable\\
        9 & can-support\\
        13 & can-contain\\
        14 & supportable\\
        \hline
        \end{tabular}
    }
    }
    }
    \vfill
    \hspace{92mm}
    \subfigure{
        \scalebox{0.7}{%
        \begin{tabular}{c c c}
        \hline
        \textbf{V-measure} & \textbf{Homogeneity} & \textbf{Completeness}\\
        \hline
        0.70 & 0.75 & 0.66\\
        \hline
        \end{tabular}
    }
    }

    \caption{Hierarchical clustering of 100 \emph{AGraphlets} captured from the CAD-120 dataset. (top) The hierarchy of clusters where the \textit{y}-axis corresponds to the distance and the \textit{x}-axis shows the cluster id every leaf node is assigned to; edges of the leaf nodes are colored depending on the cluster the leaf node is assigned to. (bottom left) Visualization in 3D space of the graph embeddings of the clustered data, after applying PCA. Coloring is associated with the clusters they are assigned to. (bottom right) Table mapping the clusters id with groundtruth labels.}
    \label{fig:object_hierarchies}
\end{figure}

\begin{figure}
    \centering
    \includegraphics[trim ={0mm 7mm 52mm 0mm},clip, width=1.\textwidth]{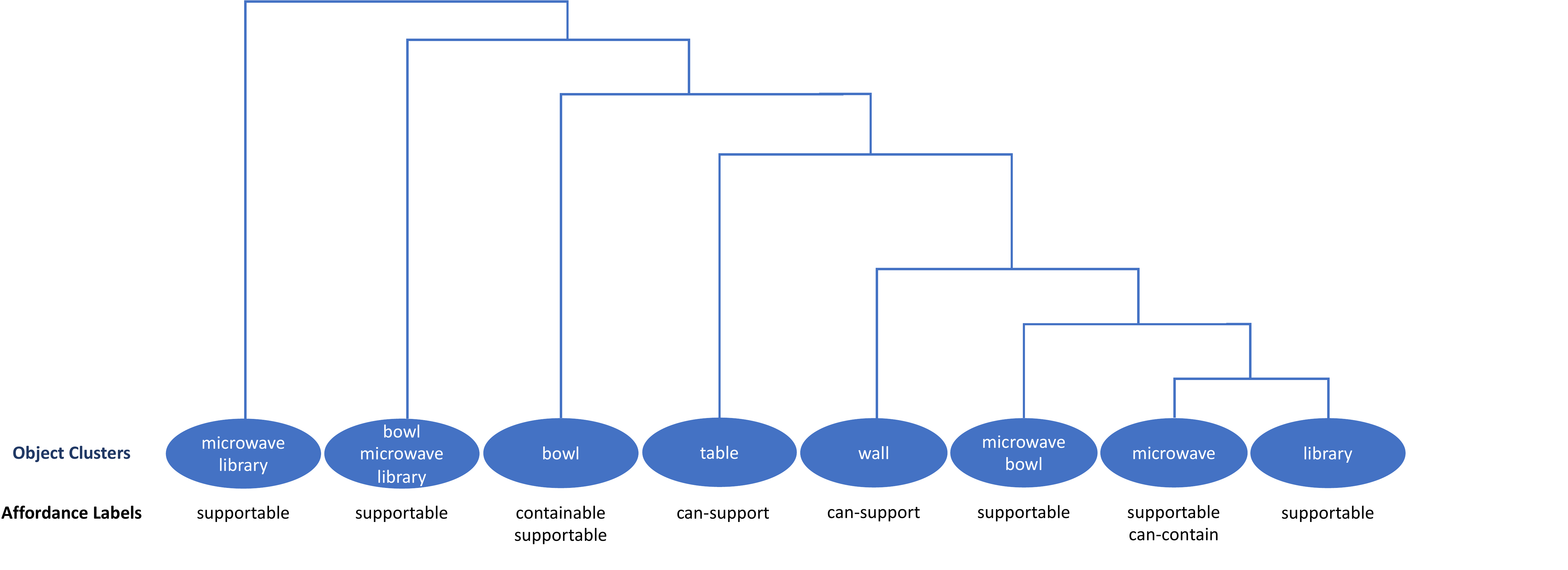}
    \caption{Clusters' hierarchy for a video from the CAD-120 dataset.}
    \label{fig:tree_cad}
\end{figure}

A representative instance of our object affordance categorization method is illustrated in Figure \ref{fig:object_hierarchies}. A subset of \emph{AGraphlets} from the test set of the CAD-120 dataset is employed in this figure as an input to the proposed methodology. Groups of similar \emph{AGraphlet} embeddings in the latent space form clusters of object affordances.
Clusters are color-coded in the presented hierarchy.

The depicted clusters denote the different kind of affordances in the present subset of a video. To facilitate better understanding of the results, by considering
the dominant ground truth affordance label in each group of data points, we are able to give an affordance name to every cluster (Fig. \ref{fig:object_hierarchies} (bottom right)).  Some clusters have been assigned the `unknown label', which denotes an affordance that is not present in the groundtruth hence, creating a fine-grained affordance hierarchy. Also, some clusters have two labels of affordances, which indicate that the objects in those clusters carry both these affordance labels.
Our clustering hierarchy of affordances takes no account of any object labels, thus a cluster is able to contain all objects that have similar interactions in reference to their \emph{AGraphlets} embeddings.

Figure \ref{fig:tree_cad} shows the clustering output of a video from the CAD-120 dataset.
Groups of objects with similar affordances form the clusters of the hierarchy. Groundtruth affordance labels are assigned to every cluster by inspecting the objects and their \emph{AGraphlets} the cluster comprise of.

An important aspect of our method, as illustrated in this figure, is that we achieve a fine-grained affordance categorization by differentiating between different kinds of the same convexity type objects, \ie the bowl and the microwave in this example. As both the bowl and the microwave are containers, the bowl plays the role of a containee. Whilst, we can also group together different convexity type objects if they carry the same interaction.

The hierarchical tree presented, is generalizable to the whole dataset. More hierarchical trees of other videos demonstrating the generalization of our approach across different datasets, thus videos, are illustrated in the appendix.


\section{Discussion} \label{sec:discussion}

\subsection{Failure Cases} \label{sec:failure_cases}
Object detections are retrieved from an off-the-shelf object detector, trained on the COCO dataset. However, some object configurations, present in the datasets used for the evaluation of this work, are out of the object detector's training distribution. Hence, false positive object detections are evident in cases where the primary objects change shape and visual appearance, \ie opening a closed microwave. In such cases, false positive relations between detected objects are captured. \Eg opening the door of a closed microwave produces a detection for the main body of the microwave object as well as a different one for its door, resulting to the presence of \emph{AGraphlets} between the door and the rest of the objects in the scene. Hence, the proposed approach is heavily reliant to the quality of object proposals.

\subsection{Limitations} \label{sec:limitations}
Our definitions of spatial interactions model spatial states of the interactive objects, \eg a cup is on the table. However, some kind of affordances are derived from interactions holding as a transition from one state to another, \eg `pourable', `able to be poured' are inferred from the transition of the state \textsf{Cont}(bowl1, liquid) to \textsf{Cont}(bowl2, liquid). Our method is limited to only detect state-based relations; hence affordances as `pourable', `able to be poured', `throwable', and `able to be thrown' are not detectable in the current pipeline.
Although the proposed framework is generic, it is currently limited to the set of detectable and defined relations.
Moreover, an enhancement of detecting interactions of more than two objects is necessary for affordances which are inferred from the interaction of multiple objects, \eg learning `stirable' requires both a liquid and a stirrer to be contained in a concave object.


\section{Conclusion} \label{sec:conclusion}
In this work, we presented a novel depth-informed qualitative representation with the ability to handle occlusions by considering depth information, and efficiently detecting relative spatial relationships between objects in the 2D image plane.
We also addressed the problem of affordance categorization by creating embeddings of graphs of human-object and object-object interactions using the \emph{graph2vec} network, and clustering these graphs in an unsupervised way, forming groups of similar affordances in the latent space. 
Moreover, we introduced a new object affordance dataset which expands the affordance domain compared to existing datasets.
Our experiments demonstrate that exploiting the proposed relations produces more accurate qualitative graphs, describing the interactions, resulting in more homogeneous and complete clusters of affordances, and higher V-measure scores.

The enhancement of the graph representations with additional object information, such as the size of the objects, is a future direction for expanding the possible affordances our method can detect.
Also, in our future work, we aim to enhance the qualitative spatial relations, by detecting multi-object relations and enabling the construction of sequence-based relations, which allow the description of interactions that occur during a series of episodes, \eg `throwing'.


\section{Acknowledgements}
The authors would like to thank Panagiotis Magkafas for his valuable assistance in re-impelementing the Aksoy et al. baseline work, used for comparison with our approach. This work has been partially supported by the European Union's Horizon 2020 research and the innovation program under grant agreement No. 825619 (AI4EU), and by an Alan Turing Institute Fellowship to the second author.
For the purpose of open access, the authors have applied a Creative Commons Attribution (CC BY) licence to any Author Accepted Manuscript version arising from this submission.

\bibliography{main}
\bibliographystyle{theapa}

\newpage
\appendix
\section{}

\subsection{LOAD Dataset}

\begin{figure*}[h!]
\centering
    \includegraphics[trim ={0mm 32mm 0mm 0mm},clip, width=1.\textwidth]{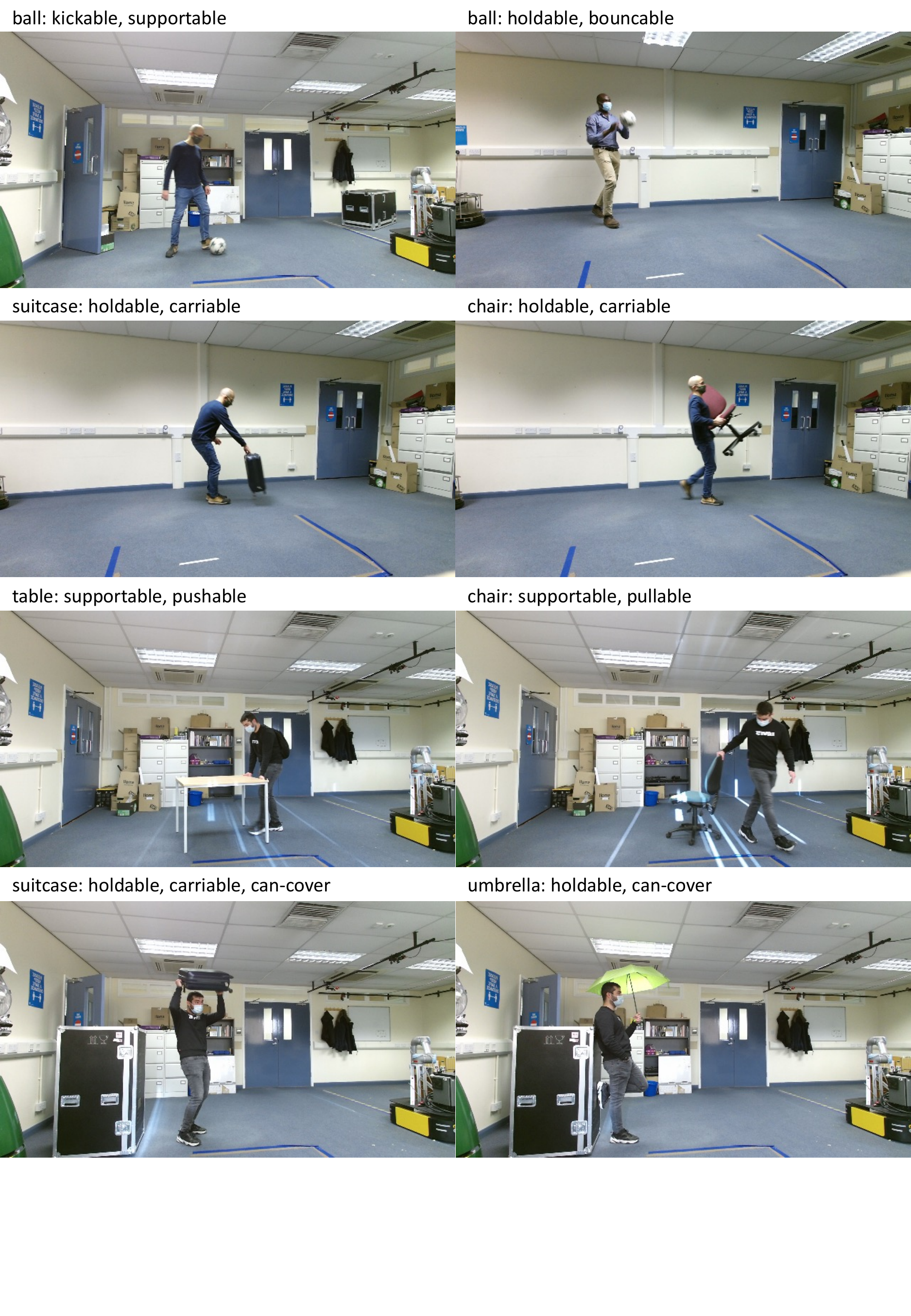}
\end{figure*}
\begin{figure*}[h!]
\centering
    \includegraphics[trim ={0mm 32mm 0mm 0mm},clip, width=1.\textwidth]{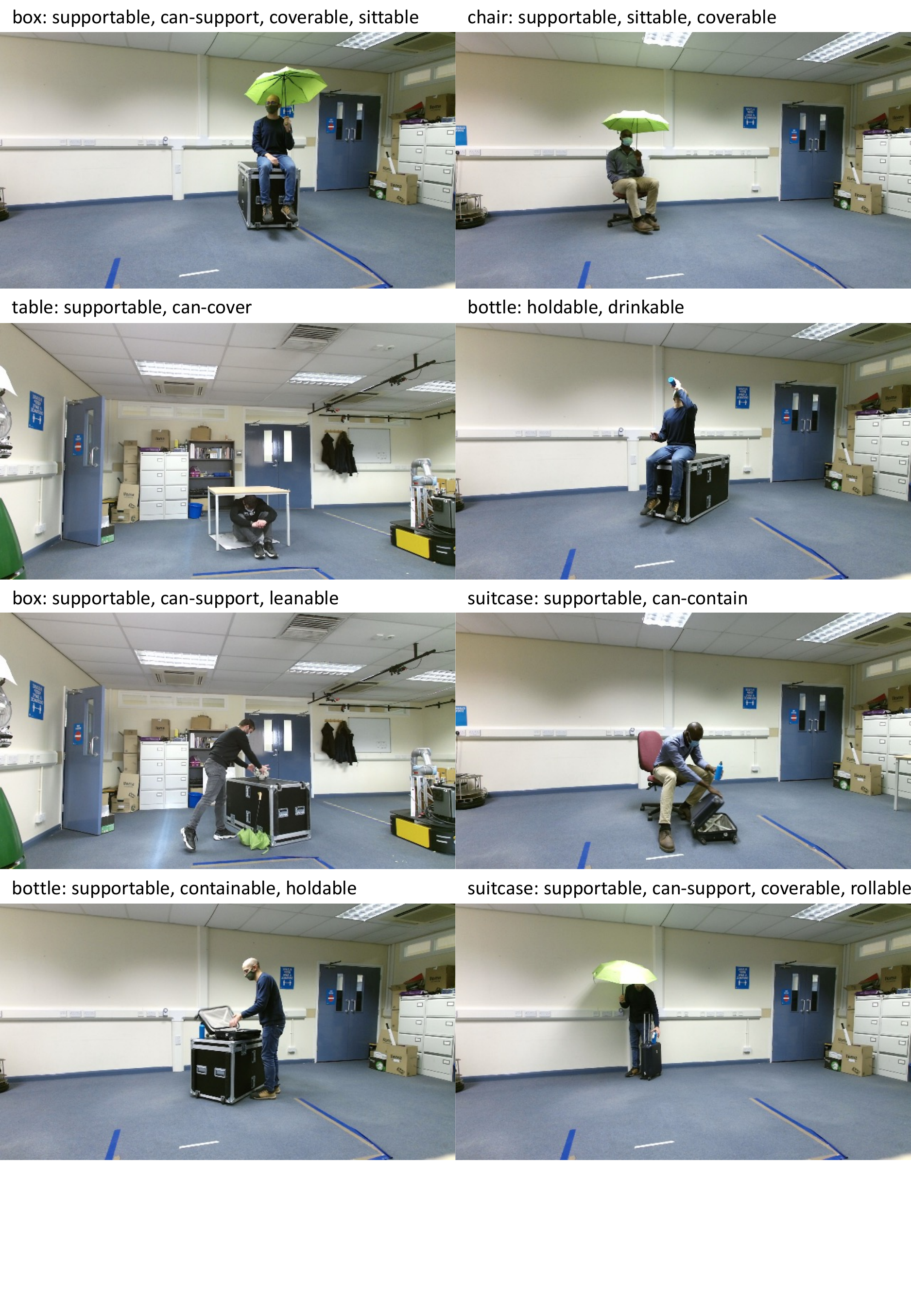}
\end{figure*}
\begin{figure*}[h!]
\centering
    \includegraphics[trim ={0mm 33mm 0mm 0mm},clip, width=1.\textwidth]{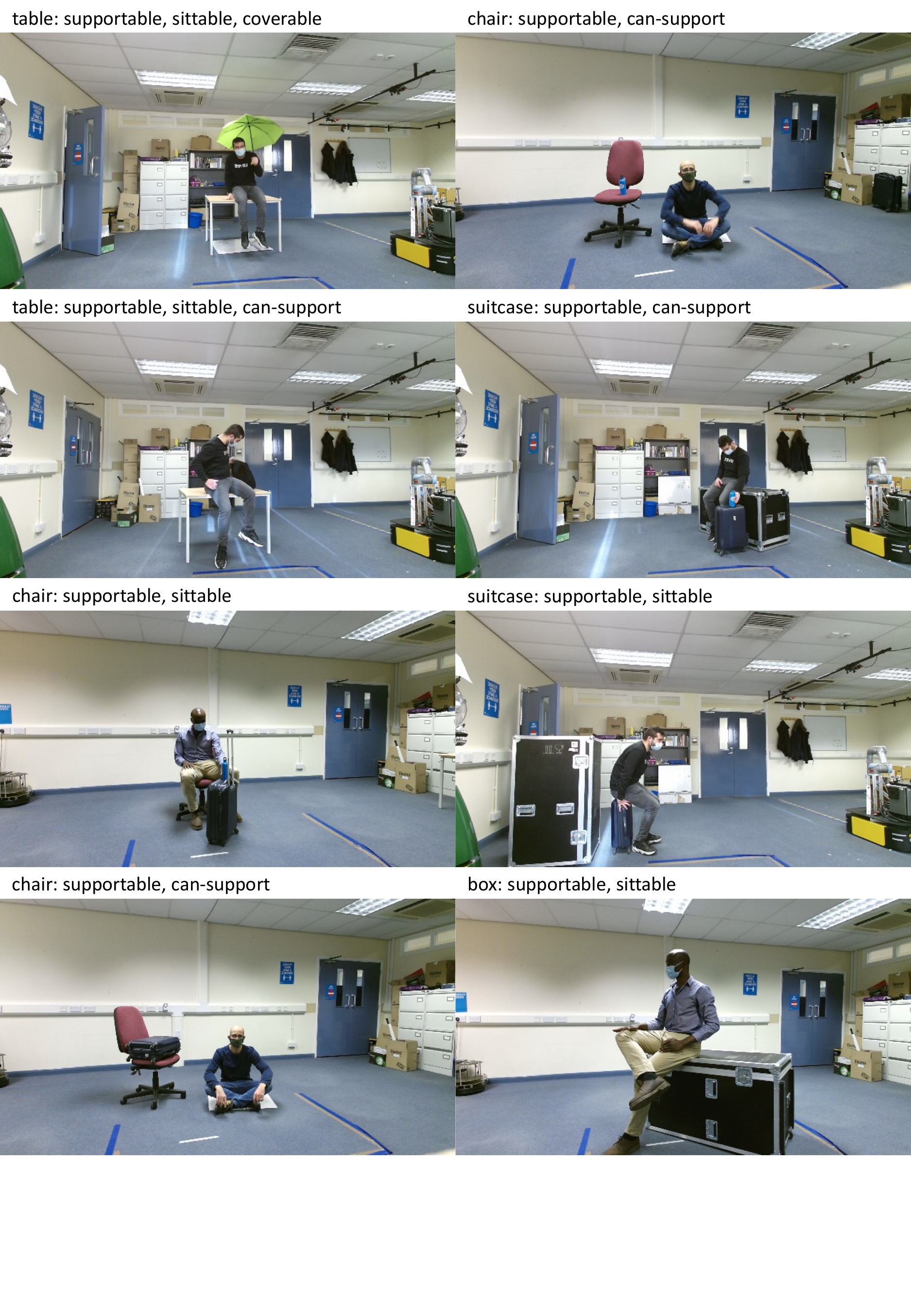}
\end{figure*}
Some sample image frames from the LOAD dataset are presented here, along with the affordance labels for one of the detected objects in every scene.

\subsection{\emph{graph2vec} Hyper-parameter Tuning} \label{sec:abl-graph2vec-hyper-parameter-tuning}

\begin{figure}
    \centering
    \includegraphics[trim ={2mm 23mm 3mm 2mm},clip, width=.9\linewidth]{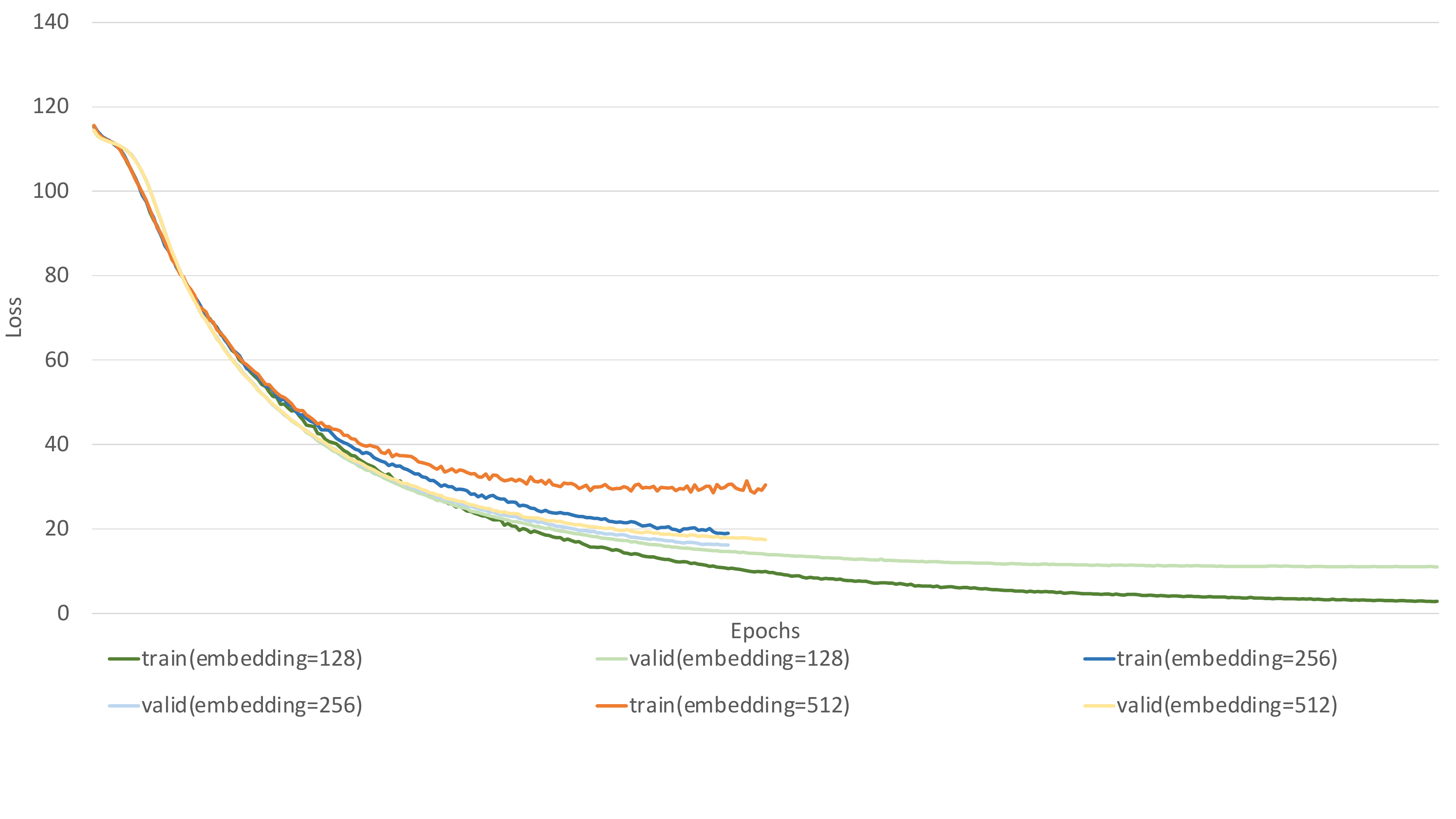}
    \caption{(best viewed in color) Training and validation loss of \emph{graph2vec} network for different embedding sizes. The experiments were done with batch size 512 and learning rate 0.5. The loss history does not update if a validation loss value greater than the latest reported has been calculated.}
    \label{fig:graph2vec_emb}
\end{figure}

\begin{figure}
    \centering
    \includegraphics[trim ={1mm 98mm 115mm 2mm},clip, width=.8\linewidth]{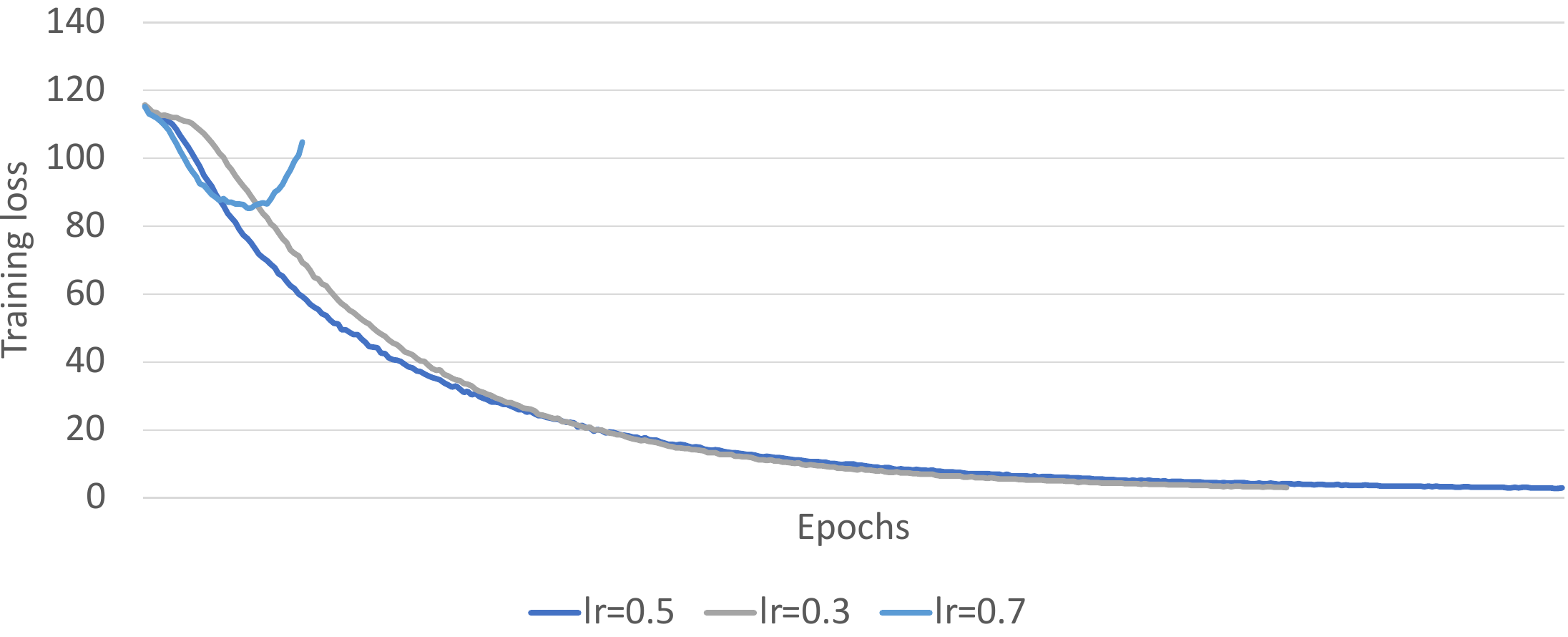}
    \caption{(best viewed in color) Training loss of \emph{graph2vec} network with embedding size 128 for different learning rates. The experiments were done with batch size 512. The training loss history does not update if a validation loss value greater than the latest reported has been calculated.}
    \label{fig:graph2vec_lr}
\end{figure}

For the selection of the embedding size we trained the \emph{graph2vec} network with the embedding size values 128, 256, 512, and concluded that an embedding of size 128 produces the smallest training and validation loss (Fig. \ref{fig:graph2vec_emb}). Batch size and learning rate remained unchanged in this experiment.
Moreover, we evaluated the output after training the network with different learning rates (0.3, 0.5, 0.7). As illustrated in Figure \ref{fig:graph2vec_lr}, the best training results were produced with learning rate 0.5. The batch and embedding sizes remained the same throughout this study. The batch size was selected through an empirical study.

\subsection{Qualitative Results}
\begin{figure}[ht]
    \centering
    \subfigure{
    \includegraphics[trim ={45mm 5mm 45mm 7mm},clip, width=.95\textwidth]{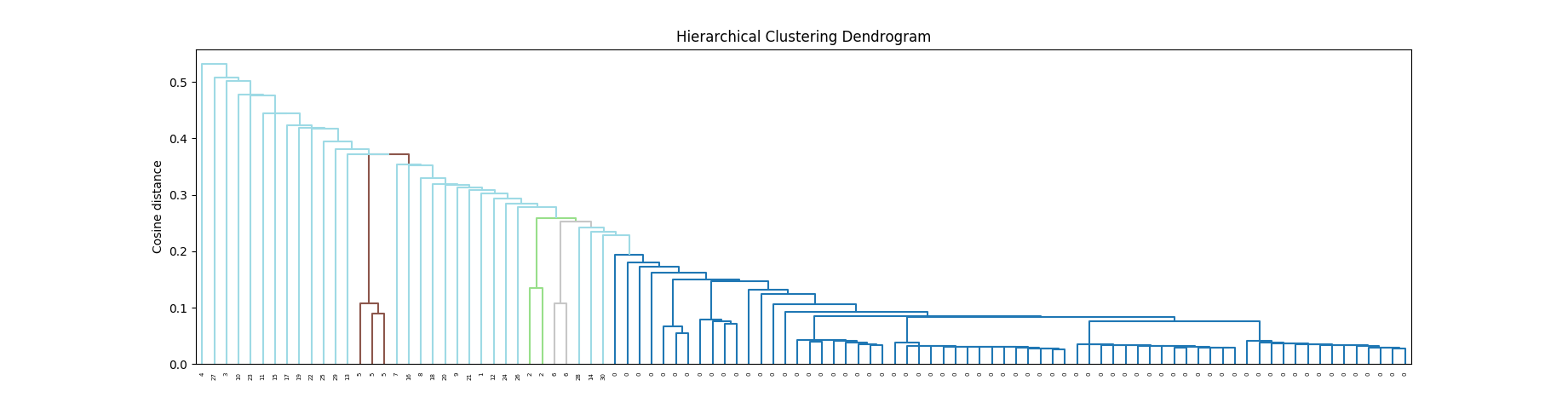}
    }
    \vfill
    \subfigure{
    \includegraphics[trim ={70mm 32mm 43mm 41mm},clip, width=.5\textwidth]{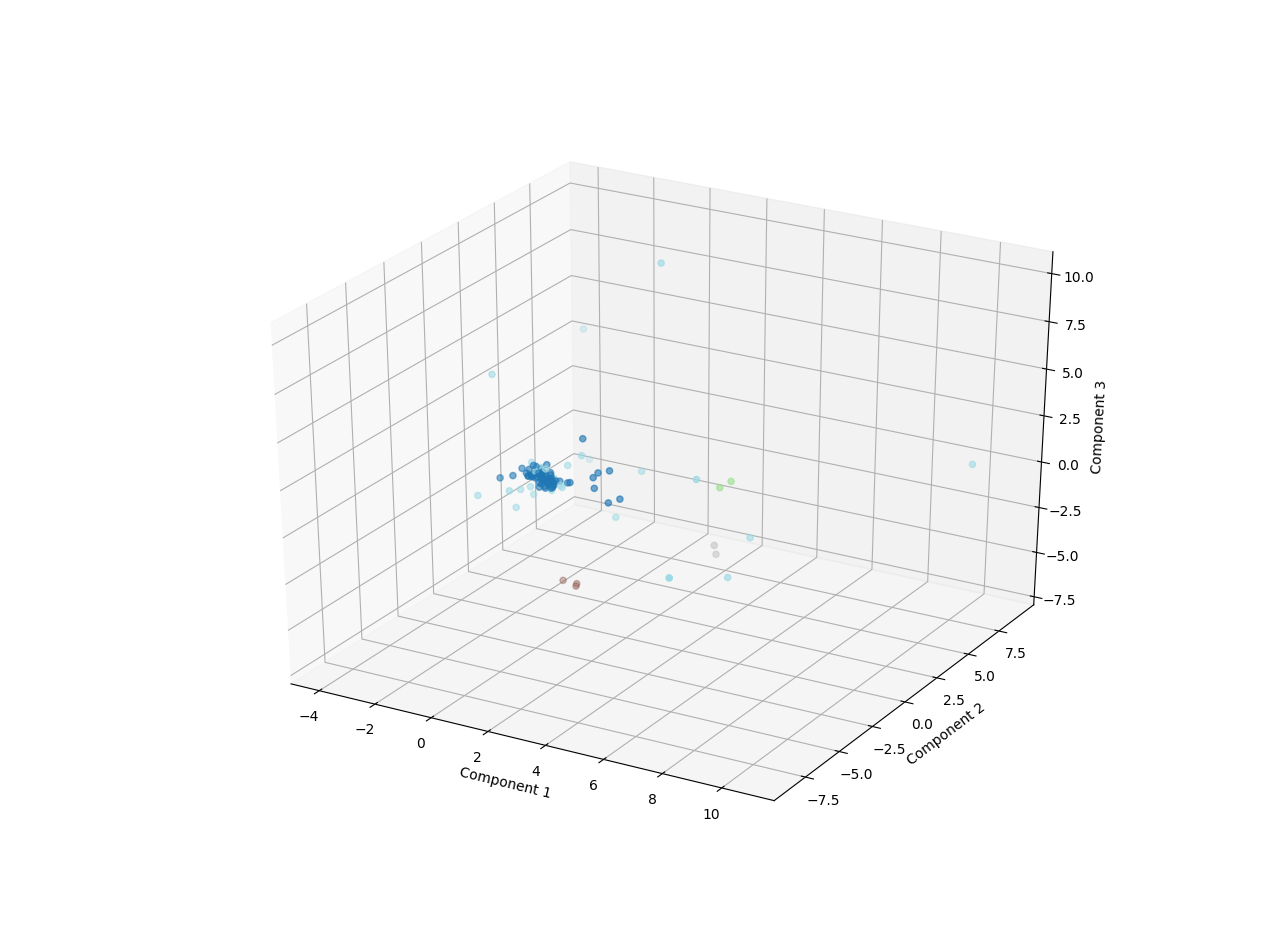}
    }
    \hfill
    \subfigure{
    \scalebox{0.7}{%
    \raisebox{5cm}{
        \begin{tabular}{||c | c||}
        \hline
        \textbf{Cluster number} & \textbf{Groundtruth label}\\
        \hline \hline
        0 & supportable\\
        2 & holdable\\
        5 & supportable\\
        6 & holdable\\
        \hline
        \end{tabular}
    }
    }
    }
    \vfill
    \hspace{92mm}
    \subfigure{
        \scalebox{0.7}{%
        \begin{tabular}{c c c}
        \hline
        \textbf{V-measure} & \textbf{Homogeneity} & \textbf{Completeness}\\
        \hline
        0.33 & 0.37 & 0.30\\
        \hline
        \end{tabular}
    }
    }
    \caption{Qualitative results from the Watch-n-Patch dataset.}
    \label{fig:qualitative_wnp}
\end{figure}

\begin{figure}[h]
    \centering
    \includegraphics[trim ={0mm 40mm 15mm 0mm},clip, width=.85\textwidth]{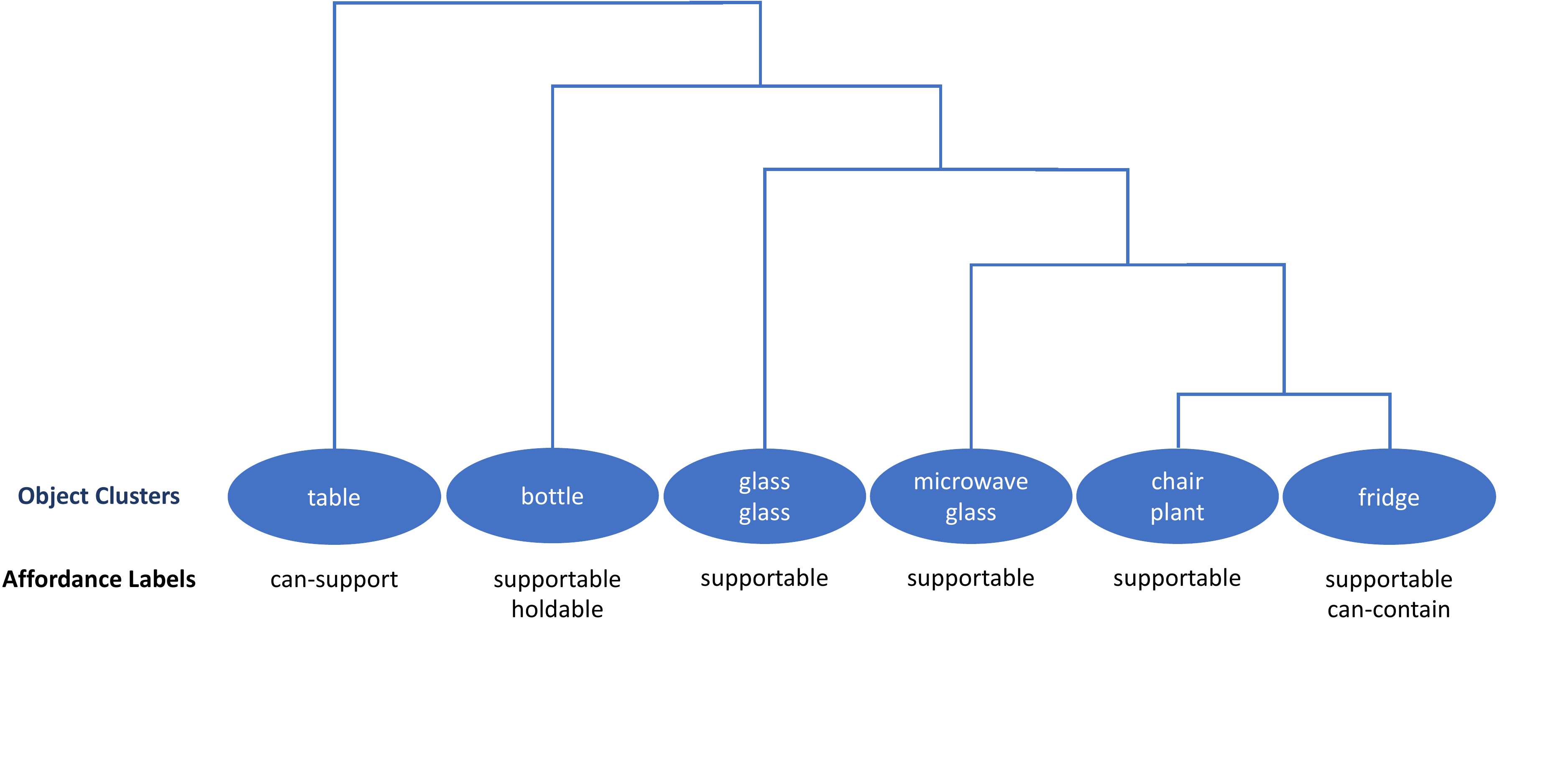}
    \caption{Clusters hierarchy for a video from the Watch-n-Patch dataset.}
    \label{fig:tree_wnp}
\end{figure}

\begin{figure}[ht]
    \centering
    \subfigure{
    \includegraphics[trim ={45mm 5mm 45mm 7mm},clip, width=.95\textwidth]{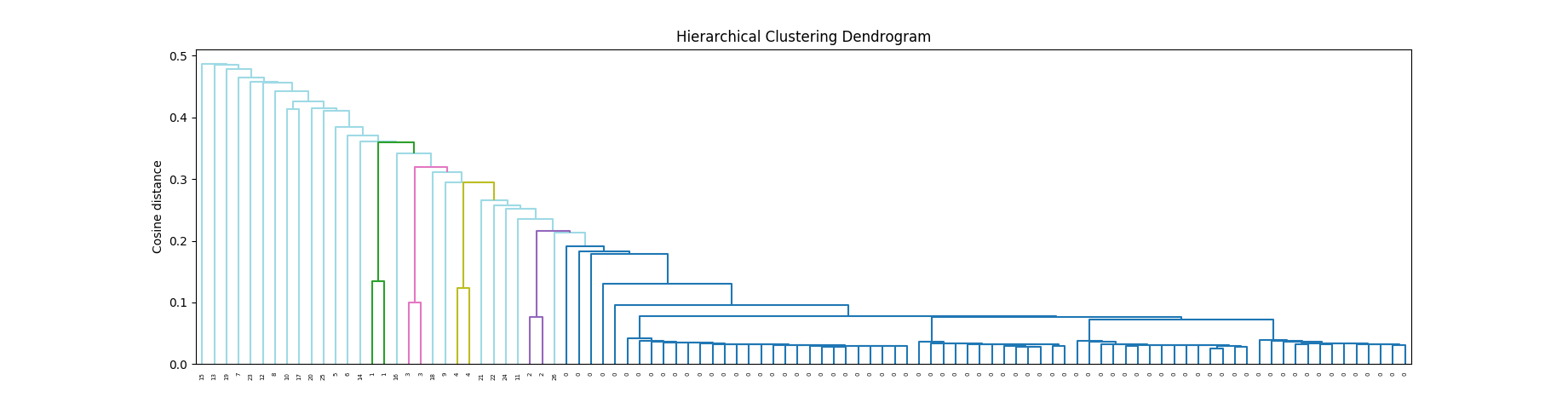}
    }
    \vfill
    \subfigure{
    \includegraphics[trim ={57mm 34mm 35mm 44mm},clip, width=.5\textwidth]{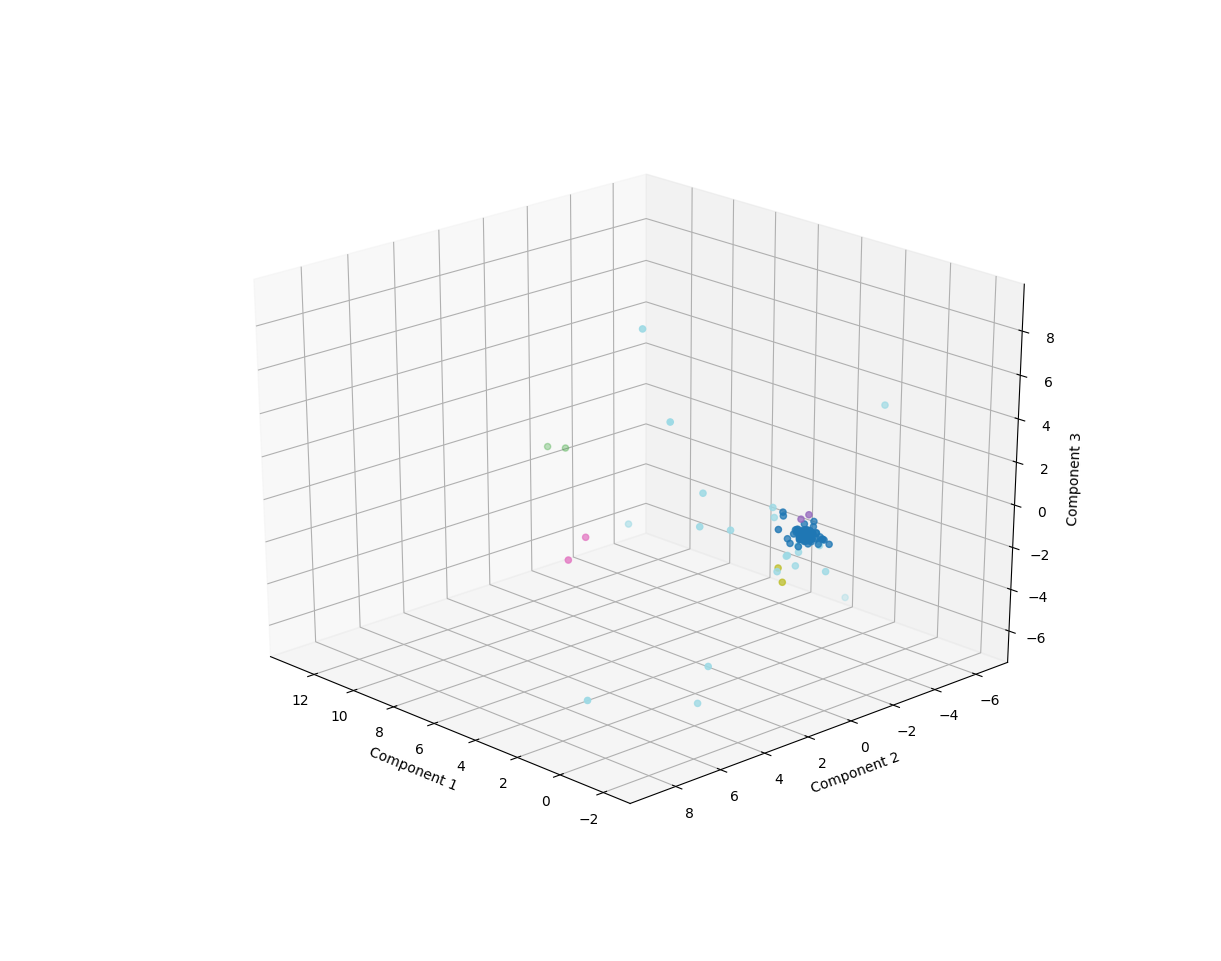}
    }
    \hfill
    \subfigure{
    \scalebox{0.7}{%
    \raisebox{5cm}{
        \begin{tabular}{||c | c||}
        \hline
        \textbf{Cluster number} & \textbf{Groundtruth label}\\
        \hline \hline
        0 & can-contain, holdable, supportable\\
        1 & unknown label\\
        2 & can-support\\
        3 & supportable\\
        4 & unkown label\\
        \hline
        \end{tabular}
    }
    }
    }
    \vfill
    \hspace{92mm}
    \subfigure{
        \scalebox{0.7}{%
        \begin{tabular}{c c c}
        \hline
        \textbf{V-measure} & \textbf{Homogeneity} & \textbf{Completeness}\\
        \hline
        0.67 & 0.63 & 0.71\\
        \hline
        \end{tabular}
    }
    }
    \caption{Qualitative results from the LOAD dataset.}
    \label{fig:qualitative_load}
\end{figure}

\begin{figure}[h]
    \centering
    \includegraphics[trim ={0mm 60mm 27mm 0mm},clip, width=.9\textwidth]{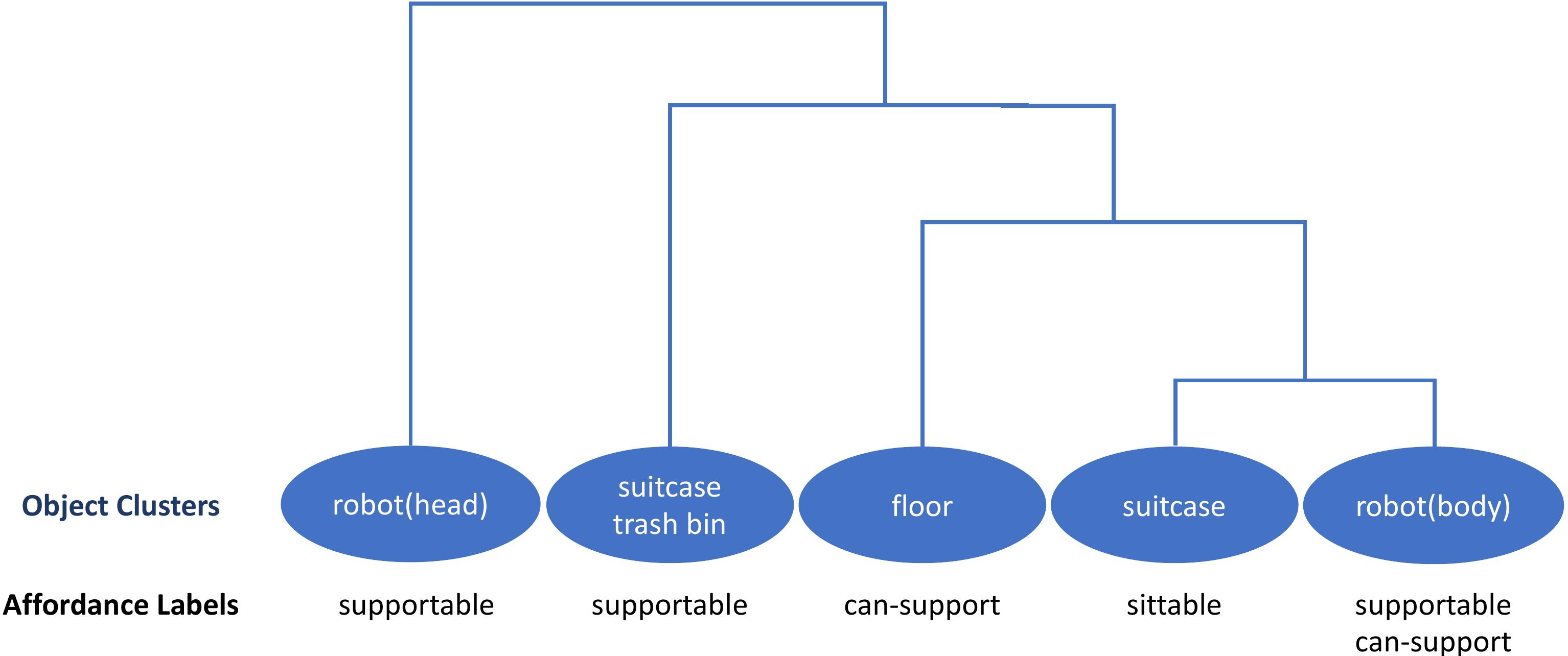}
    \caption{Clusters hierarchy for a video from the LOAD dataset.}
    \label{fig:tree_load}
\end{figure}

Further qualitative results are presented in this section (Fig. \ref{fig:qualitative_wnp}, Fig. \ref{fig:qualitative_load}). We also, present the cluster hierarchy formed, illustrating groups of object affordances (Fig. \ref{fig:tree_wnp}, Fig. \ref{fig:tree_load}). Every cluster forms a group of objects that share the same affordance. Some clusters comprise a single object, however with multiple affordances.

\end{document}